\documentclass[10pt,twocolumn,letterpaper]{article}

\usepackage{cvpr}
\usepackage{times}
\usepackage{epsfig}
\usepackage{graphicx}
\usepackage{amsmath}
\usepackage{amssymb}
\usepackage{mathtools}
\usepackage{multirow}
\usepackage{makecell}
\usepackage{pifont}
\usepackage{subcaption}
\newcommand{\cmark}{\ding{51}}
\newcommand{\xmark}{\ding{55}}
\usepackage{multicol}
\usepackage{pstricks}

\usepackage{multirow}
\usepackage{xcolor}
\newtheorem{theorem}{Theorem}

\def\ie{\textit{i.e.}, }

\usepackage[page]{appendix}

\newcommand{\tr}{\mbox{$^{\top}$}}

\def\R{{\rm I} \! {\rm R}}

\newcommand{\eq}[1]{Eq.~(\ref{eq:#1})}

\newcommand{\SKIP}[1]{} 

\newcommand{\vv}[2]
   {\mbox{$
      \left(
         \begin{array}{c}
         #1 \\ #2
         \end{array}
      \right )
   $}}
\newcommand{\mbegin} {\left [ \begin{array}}

\newcommand{\mend}   {\end{array} \right ]}

\newcommand{\detbegin} {\left | \begin{array}}
\newcommand{\detend}   {\end{array} \right |}

\newcommand{\vbegin} {\left ( \begin{array}{c}}

\newcommand{\vend} {\end{array}\right )}

\def\squareforqed{\hbox{\rlap{$\sqcap$}$\sqcup$}}

\def\qed{\ifmmode\squareforqed\else{\unskip\nobreak\hfil
	\penalty50\hskip1em\null\nobreak\hfil\squareforqed
	\parfillskip=0pt\finalhyphendemerits=0\endgraf}\fi}

\def\vec#1{\mathchoice%
	{\mbox{\bf $\displaystyle\bf#1$}}
	{\mbox{\bf $\textstyle\bf#1$}}
	{\mbox{\bf $\scriptstyle\bf#1$}}
	{\mbox{\bf $\scriptscriptstyle\bf#1$}}}
\def\vv#1{\protect\vec #1}

\newcommand{\showeqnlabel}{
	\hbox to 0pt{\quad\quad\relax\fbox{\scriptsize\rm\eqnlblx}%
	\gdef\eqnlblx{xxxx}}} \newcommand{\eqnlblx}{}

\def\@eqnnum{\rm (\theequation)\showeqnlabel}

\newcommand{\nofig}[1]{\centerline{\bf Figure here}}

\def\mat#1{\mathchoice{\mbox{\bf$\displaystyle\tt#1$}}
	{\mbox{\bf$\textstyle\tt#1$}}
	{\mbox{\bf$\scriptstyle\tt#1$}}
	{\mbox{\bf$\scriptscriptstyle\tt#1$}}}
\def\m#1{\protect\mat #1}

\newcommand{\zw}[1]{{\blue Zw: #1}}

\def\myred#1{{\color{red}{#1}}}

\usepackage[pagebackref=true,breaklinks=true,letterpaper=true,colorlinks,bookmarks=false]{hyperref}
\hypersetup{colorlinks=true, linkcolor=red, filecolor=magenta, urlcolor=blue}

\cvprfinalcopy 

\def\cvprPaperID{2166} 

\begin{document}

\newif\ifsupp
\newif\ifmain
\newif\ifarxiv
\newif\ifcomment

\commentfalse
\suppfalse
\mainfalse
\arxivtrue

\ifsupp
    \title{Refining Semantic Segmentation with Superpixels using 
Transparent \\ Initialization and Sparse Encoder\\
    -- Supplementary Material --}
\else
    \title{Refining Semantic Segmentation with Superpixels using 
Transparent \\ Initialization and Sparse Encoder}
\fi

\author{Zhiwei Xu$^{1,2}$ \quad Thalaiyasingam Ajanthan$^1$ \quad \quad Richard Hartley$^1$\\
$^1$Australian National University and Australian Centre for Robotic Vision\\
$^2$Data61, CSIRO, Australia\\
{\tt\small \{zhiwei.xu,thalaiyasingam.ajanthan,richard.hartley\}@anu.edu.au}
}

\ifsupp
    \maketitle
\else
    \twocolumn[{
    \renewcommand\twocolumn[1][]{#1}
    \maketitle
    \begin{center}
        \includegraphics[width=\textwidth]{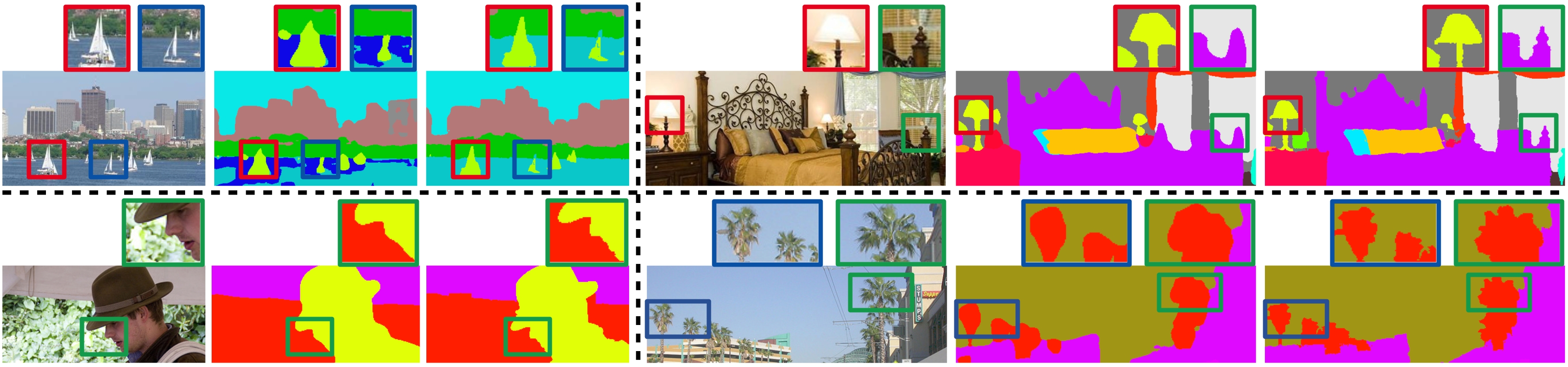}
        \vspace{-1.2\baselineskip}
        \captionof{figure}{\em\small Edge comparison with the state-of-the-art, that is ResNeSt200 on ADE20K (top row) and
        ResNeSt269 on PASCAL Context (bottom row).
        \textbf{Left: RGB, middle: state-of-the-art, right: ours}.
        Ours has a better alignment with object edges than the state-of-the-art.}
        \label{fig:demo}
    \end{center}
    }]
\fi

\ifsupp
    \setcounter{equation}{14}
    \setcounter{figure}{8}
    \setcounter{table}{6}
    \setcounter{theorem}{1}

    \appendix
    This provides additional details about transparent layers, giving more
details of the initialization and structure of the layers.
In particular, it shows how the approach may be used for any activation functions,
not only forms of ReLU that were discussed in the main paper.
This also speculates about the application of these ideas to form transparent
convolutional layers.
We now describe the ideas at a slower pace than is possible in the main paper,
sometimes using different notation.


\section{Introduction for Warmup}
A common technique is to extend an existing network by the addition of further 
affine layers (classification layers) before the loss layer.  These layers may then
be trained separately or the whole extended network can be trained.
We refer to the original (unextended) network as the base-network, and 
the network extended by some additional layers as the extended network.
We assume that the extension layers consist of fully-connected layers,
including offset, followed by an activation layer, such as a ReLU.  These extension
layers are inserted before the loss-layer.

Assuming that the base layer is trained to attain a low value of the loss, we wish
to insert the extension layers into the network without increasing the loss of
the network, so that additional training may continue to decrease the value
of the loss-function.  However, if the extension layers are initialized with
random parameters, it will result in an increase in the loss of the network,
so that the extension layers, or the whole network needs to be trained from
scratch.  

\paragraph{Simple transparent layers.}
This problem has also been addressed in previous work of Goodfellow
\cite{net2net}.
One very simple method to implement a transparent layer is
to let the matrix of layer parameters be the identity matrix.  This has the disadvantage, however,
that all the parameters are either $0$ or $1$, which fails to introduce
sufficient randomness into the system, which would perhaps be desirable.
In addition, if the function $f$ is represented by the identity mapping, then
many of the parameters are equivalent, perhaps also an undesirable feature.
We wish to allow the parameters of the affine transformation to be chosen randomly so as to mix things up a bit.

It is also not clear to do in the case where $f:\R^m \rightarrow \R^n$,
where $m \ne n$, for then there is no such thing as an identity matrix.
We wish to allow the possibility that the output and input of the extension
layers are not equal.

\section{Affine Layers}
We are interested in layers in a Neural Network (NN) that implement
an affine transformation on the input, that is, a linear transformation
followed by an offset.  For the present, we ignore any
non-linear activatation that might be applied to the output of the layer.

The most obvious example of such a layer is a fully-connected layer, and
that will be our main focus.  However, the same idea could be applied
to handle convolution layers.  We raise this possibility and make a few comments
on this later. Any development of this idea is left to further work.

An affine layer, in the following discussion, is seen as composed of a linear transform,
followed by an offset and then a non-linear activation.  The linear transform,
followed by offset performs an affine transformation on the data, which 
will be written as%
\footnote{
The notation in this exposition differs slightly, mainly in choice of fonts, from 
the main paper. Here, we deonte a matrix by $\vv A$, an affine transformation
by $\m A$ and the matrix that expresses an affine transformation in terms of
homogeneous coordinates by $\tilde{\vv A}$.  Homogeneous quantities in general
(such as homogeneous vectors are denoted with a tilde).
}
\begin{equation}
f(\vv x) = \vv x \vv A + \vv b
\end{equation}
where $f:\R^m \rightarrow \R^n$. 
In this note, the convention is that vectors, such as $\vv x$ are {\rm row vectors},
which is contrary to a convention common in computer vision literature
that vectors represent column vectors.  The present notation is common
in computer graphics literature, however.

This mapping satisfies the condition
\begin{equation}
f(\lambda \vv x) + f((1-\lambda) \vv x) = f(\vv x)
\end{equation}
and it may be represented by
matrix multiplication on the homogenized vector $\tilde{x}$, according to
\begin{equation}
\tilde x \mapsto  \tilde x \tilde{\vv A}
\end{equation}
where the matrix $\tilde {\vv A}$ is given by
\begin{equation}
\tilde {\vv A} = \mbegin{cc}
\vv A & \vv 0\tr \\
\vv b  & 1
\mend  ~.
\end{equation}

For future reference, the matrix that performs the inverse affine transformation
is equal to
\begin{equation}
\tilde {\vv A}^{-1} = \mbegin{cc}
\vv A^{-1} & \vv 0\tr \\
-\vv b \vv A^{-1} & 1
\mend  ~.
\end{equation}

This works if $\vv A$ is a square matrix and has an inverse.
More generally, in the case where $\vv A$ is of dimension $m \times n$,
with $m < n$, it is possible that $\vv A$ has a right-inverse, which is
a matrix, denoted by $\vv A^R$ such that $\vv A \vv A^R = \vv I$.
In this case, a right inverse for $\tilde {\vv A}$ is given by
\begin{equation}
\tilde {\vv A}^{R} = \mbegin{cc}
\vv A^{R} & \vv 0\tr \\
-\vv b \vv A^{R} & 1
\mend  ~.
\end{equation}
where $\vv A^R$ is the right-inverse of matrix $\vv A$.

An affine transform maps $\R^m$ onto an affine subspace of $\R^n$,
and the {\em rank} of the affine transform may be defined as the dimension
of the affine space that is the image of this transform.  Such a transformation
will be said to be of {\em full rank} if its rank is equal to $m$, the
dimension of input space.

From here on, we shall use the symbol $\m A$ to represent an affine
transformation, writing $\vv x \m A$.
The symbol $\m A$ represents the
transformation itself, not the matrix (in this case $\tilde {\vv A}$) that implements it.
{\bf Caution:} to reemphasize this, the symbols $\vv A$ and $\m A$ do not represent
the same thing.  In particular $\vv A$ is a matrix and $\m A$ is an affine transform,
which is equivalent to matrix multiplication by a matrix $\vv A$ and offset
by a vector $\vv b$, so that $\vv x \m A$ is short-hand notation for $\vv x \vv A + \vv b$.

\paragraph{Interlude: right inverse of a matrix.}
\label{sec:interlude}
Let $\vv A$ be an $m \times n$ matrix with $m \le n$.
An $n\times m$ matrix $\vv A^R$ is the right-inverse of $\vv A$ if $\vv A \vv A^R = \vv I_{m\times m}$.
Such a matrix exists if and only
if $\vv A$ has rank $m$ equal to its row-dimension.  
This cannot happen if $m > n$, where a right-inverse cannot exist.

In this case, $\vv A^R$ is given by the formula
\begin{equation}
\vv A^R = \vv A\tr (\vv A \vv A\tr)^{-1}\ ,
\end{equation}
where the condition $\vv A \vv A^R = \vv I$ is easily verified.
This formula 
relies on the fact that $\vv A \vv A\tr$ has an inverse, which is ensured because
the rank of $\vv A$ is $m$.

An alternative procedure is to take the Singular Value Decomposition (SVD)
$\vv A = \vv U \vv D \vv V\tr$, where $\vv V\tr$ has dimension $m\times n$
and $\vv V\tr \vv V = \vv I_{m\times m}$.
The matrix $\vv D$ (of dimension $m\times m$)
is non-singular (since $\vv A$ has rank $m$).  Then, the right-inverse is
given by
\begin{equation}
\vv A^R = \vv V \vv D^{-1} \vv U\tr
\end{equation}
as is easily verified.

\section{Sequences of Layers without Activation}
For the time being, we consider the unrealistic case that the activation layer
is missing from the affine layers.  In this case, each affine layer performs
an affine transformation.  A sequence of transformations 
$\m A_1 \m A_2 \ldots \m A_k$ may be applied to an input $\vv x$, giving
\begin{equation}
\vv x \mapsto \vv x \m A_1 \m A_2 \ldots \m A_k ~.
\end{equation}
If we choose $\m A_k$ to be a right-inverse of $\m A_1 \m A_2 \ldots \m A_{k-1}$
then we have
\begin{equation}
\m A_1 \m A_2 \ldots \m A_k = \m A_1 \ldots \m A_{k-1} (\m A_1 \ldots \m A_{k-1})^R = I ~,
\end{equation}
the identity mapping.

The condition for this right inverse to exist is that $\m A_1 \m A_2 \ldots \m A_{n-1}$ should
be of full rank.  A necessary condition for this to happen is as follow.
Let $\m A_i$ represent an affine transformation $\m A_i : \R^{m_{i-1}} \rightarrow \R^{m_i}$.
The first and last spaces in this sequence
are $\R^{m_0}$ and $\R^{m_k}$,
so $m_0 = m$ and $m_k = n$. 
Then $\m A_1 \ldots \m A_{k-1}$ has a right inverse
only if $m_i \ge m_0$ for all $i=1, \ldots, k-1$.  In other words, a necessary condition
for the right inverse to exist is that the dimensions of all the intermediate
spaces $\R^{m_1}, \ldots ,\R^{m_{k-1}}$ should be at least equal to $m_0$.
It is not hard to see that {\em generically}, this is also a sufficient condition,
meaning that it is true for almost all sequences of transformations.%
\footnote{The terms {\em almost all, almost always, etc. } are intended in the standard
mathematical sense, meaning that the set of cases for which the relevant condition
fails to be true has measure or probability zero.  For instance the set of $m\times n$,
with $m \le n$ that do not have full rank is a set of measure $0$ in the set of all
such matrices, so one can state that almost all such matrices have full rank.

As another example, a matrix chosen at random (which implies that numbers
are chosen from some probability distribution such as a normal distribution) will
be of full rank with probability $1$, hence almost always.
}

Without being too formal here, we can state that this condition will hold
{\em generically} if all the transformations are chosen at random.

\begin{theorem}
Let a sequence of affine transformations $\m A_i: \R^{m_{i-1}} \rightarrow \R^{m_i}$,
for $i=1, \ldots, k-1$
be chosen at random.  Then $\m A_1 \ldots \m A_{k-1}$ almost surely has a right inverse if and
only if $m_i \ge m_0$ for $i=1, \ldots, k-1$.
\end{theorem}
For practical purposes it is safe to proceed as if this result holds always (and not
just almost always), since the probability that $\m A_1\ldots \m A_{k-1}$ does not
have an inverse is vanishingly small (in fact zero).

Therefore, the strategy for selecting a set of parameters for a sequence of
affine layers may be described.  For the sequence of layers to represent an
identity transform, we need that the input and output space have the same
dimension, namely $m_0 = m_k$.  Then
\begin{enumerate}
\vspace{-0.3\baselineskip}
\item Select intermediate dimensions $m_1, \ldots, m_{k-1}$ such that
$m_i \ge m_0$ for all $i=1, \ldots, k-1$.
\vspace{-0.3\baselineskip}
\item Define random affine transforms $\m A_i$ by selecting the entries of 
matrices $\mathbf A_i$ and vectors $\vv b_i$ randomly, using a suitable random number
generator, for instance by a zero-mean normal (Gaussian) distribution.
\vspace{-0.3\baselineskip}
\item Compute the composition $\m A_1 \m A_2 \ldots \m A_{k-1}$ by matrix multiplication,
and take its right-inverse.
\vspace{-0.3\baselineskip}
\item Set $\m A_k$ to equal $(\m A_1 \m A_2 \ldots \m A_{k-1})^R$.
\vspace{-0.3\baselineskip}
\end{enumerate}
The resulting product $\m A_1 \m A_2 \ldots \m A_k$ is the identity affine transformation.

\section{Gaussian Random Matrices}
\label{sec:gaussian}

Since we are using matrices initialized from a normal distribution,
we give some properties of such matrices.
The purpose is to ensure that the concatenation of affine transformations
does not cause explosion of the entries of the product
$\m A_1 \m A_2 \ldots \m A_{k-1}$.  If this is not done properly, then this
product will have very large entries and so will any value
of $x_0 \m A_1 \m A_2 \ldots \m A_{k-1}$.

The solution is to make the matrices $\vv A_i$ appearing as the linear-transform
part of $\vv A_i$ as close to orthogonal as possible.  As an additional advantage,
the entries of $\m A_k = (\m A_1 \ldots \m A_{k-1})^R$ will be approximately of the
same order as the entries of each  of the other $\m A_i$.
It will turn out that the entries of each $\m A_i$ should be chosen from 
a Gaussian distribution with variance $1/m$ where $m$ is its row-dimension.

We are assuming here that the dimensions of the matrix are large.  In addition,
if the matrix is not square, by saying it is orthogonal we mean that both its
rows (and similarly it columns) are orthogonal vectors, all of the same length
(a different length for the row and column vectors, of course).
One may mistakenly assume that by randomly selecting each entry of a matrix randomly
one obtains a random matrix in some vague sense.  In reality, one obtains 
an (nearly) orthogonal matrix.

To see this, we first see that two random vectors from distribution $\mathcal{D}$
are almost orthogonal.
Let $X$ and $Y$ be two i.i.d random variables in
$\mathcal{N}(0, \sigma^2)$, we have
\begin{equation}
E[XY] = E[X] E[Y] = 0
\label{eq:inner_product}
\end{equation}
and
\begin{equation}
E[(XY)^2] = E[X^2] E[Y^2]\ .
\end{equation}
Then,
\begin{equation}
\begin{aligned}
\text{var}(XY)
&= E[(XY)^2] - E^2[XY] \\
&= E[X^2]~E[Y^2] \\
&=\left( E[X^2] - E^2[X])~(E[Y^2] - E^2[Y] \right) \\
&=\text{var}(X)~\text{var}(Y)\ .
\end{aligned}
\label{eq:var_xy}
\end{equation}
Now, given a column vector $\mathbf{v} \in \mathbb{R}^m$ with entries
chosen from distribution $\mathcal{D}$, we have
\begin{equation}
\begin{aligned}
\text{var}\left(\sum^{m}_{i=1} X_i Y_i\right)
= m\text{var}\left(X_i Y_i\right)
= m\text{var}^2(X)
= m\sigma^4,
\end{aligned}
\end{equation}
where $X_i$ and $Y_i$ are independent entries in two of such column
vectors respectively.
Then, the \textit{expected squared length} of such a vector $\mathbf{v}$ is
$mE[X^2]=m\sigma^2$.
If we choose $\sigma^2 = 1/m$, $\mathbf{v}$ has the expected squared
length equal to 1.
This satisfies the attribute of an orthogonal matrix, $\mathbf v^T
\mathbf v=\mathbf{1}$.
Then, Eq.~\eqref{eq:var_xy} follows
\begin{equation}
\text{var}\left(\sum^{m}_{i=1} X_i Y_i\right)
= m \sigma^4 = \frac{1}{m}\ .
\end{equation}
Hence, two $m$-length vectors randomly chosen from
$\mathcal{N}(0, 1/m)$ will
have expected squared length 1 and expected inner-product 0 (by
Eq.~\eqref{eq:inner_product}) with variance of the inner product as $1/m$.

On the other hand, the \textit{variance of the square length} of the
vector is
\begin{equation}
\begin{aligned}
\text{var}\left(\sum^{m}_{i=1}X_i^2\right)
= m \text{var}(X^2)
= \frac{2}{m}\ ,
\end{aligned}
\end{equation}
where $\text{var}(X^2) = 2 \sigma^4$ is obtained by $E(X^4) - E^2(X)$ with
$E^2(X^2) = \sigma^4$ and
\begin{equation}
\begin{aligned}
&E(X^4) \\
= &\frac{\int {X^4 \exp \left( {-\frac{X^2}{2\sigma^2}} \right) dX}}{\int {\exp \left( {-\frac{X^2}{2\sigma^2}} \right) dX}} \\
= &\frac{-\sigma^2 X^3 \left. \exp \left(-\frac{X^2}{2\sigma^2} \right) \right\vert^{+\propto}_{-\propto} + \sigma^2 \int {\exp \left( {-\frac{X^2}{2\sigma^2}} \right) dX^3}}{\int {\exp \left( {-\frac{X^2}{2\sigma^2}} \right) dX}} \\
= & \frac{3 \sigma^2 \int {X^2 \exp \left( {-\frac{X^2}{2\sigma^2}} \right) dX}}{\int {\exp \left( {-\frac{X^2}{2\sigma^2}} \right) dX}} \\
= & \frac{3\sigma^2 (-\sigma^2) \left( X \left. \exp \left( {-\frac{X^2}{2\sigma^2}} \right) \right\vert^{+\propto}_{-\propto} - \int {\exp \left( {-\frac{X^2}{2\sigma^2}} \right) dX} \right)}{\int {\exp \left( {-\frac{X^2}{2\sigma^2}} \right) dX}} \\
= & 3 \sigma^4\ .
\end{aligned}
\end{equation}

To this end, it shows
\begin{theorem}
If entries of an $m \times n$ matrix are chosen
from a zero-mean distribution with variance $\sigma^2=1/m$, then the
column vectors have expected squared length 1 with variance $2/m$
and the expected inner product of each two columns is 0 with variance
$1/m$.
\end{theorem}

As $m$ increases, the matrix approximates more and more an orthogonal
matrix.
Corresponding to Fig.~\myred{4} in the main paper,
weights $A_i$ and bias $b_i$ can be initialized by

\begin{enumerate}
\vspace{-0.3\baselineskip}
\item Choose the dimension of $i$-th layer filter as $m_{i-1} \times m_i$, for
$i = 1, ..., k-1$, where $m_i \ge m_0 = m_k$.
\vspace{-0.3\baselineskip}
\item Define a random affine transformation $\m A_i$ by selecting its weight matrix $\vv A_i$ from a $\mathcal{N}(0, 1/m_i)$ Gaussian distribution
and its bias vector $\vv b_i$ from a $\mathcal{N}(0, 1)$ Gaussian distribution.
\vspace{-0.3\baselineskip}
\item Initialize $\m A_k$ with the right inverse of $\m A_1 \m A_2...\m A_{k-1}$.
\end{enumerate}
Again, \textbf{note} that $\m A_i$ is an affine transformation (also used as a layer) where $\vv A_i$ is the weight matrix of $\m A_i$.
Then, the above initialization will lead to an identity mapping as $\m A_1 \m A_2...\m A_k=\mathbf{1}$.

\section{Getting Past the Activation Layer}
\label{sec:representation}
We concentrate first on the case where each affine transform is followed by 
a ReLU activation.  To do this, we will need to double the size of the intermediate
layers, as will be seen next.
We assume that affine transforms $\m A_i$ are given,
let $\vv x = \vv x_0$ be the input of $\m A_1$ and define $\vv x_i = \vv x_{i-1} \m A_i$
for $i=1, \dots, k$, the output of the $i$-th affine transform.

In the course of the following discussion, we shall be
defining new affine transforms $\m A_i'$ and layers
represented by $\m A_i' \sigma$, namely an affine transform followed by
an activation $\sigma$.   Once more let $\vv x_0'$ be the input of the first
layer, $\vv x'_0 = \vv x_0$, and define $\vv x'_i = \vv x'_{i-1} \m A'_i \sigma$, the result
of the affine transformation followed by the activation.  
Quantities without primes ($\m A_i$ and $\vv x_i'$) belong
to the original sequence of affine transforms, whereas those with primes
($\m A'_i$ and $\vv x_i'$) belong to the sequence, with activation, being constructed).

The first layer will be modified as follows.
Suppose that $\m A_1$ is represented in matrix form as
\begin{equation}
\vv x_0 \m A_1 = (\vv x_0, 1)  \mbegin{c}
\vv A_1 \\
\vv b_1 
\mend  
= \vv x_0 \vv A_1 + \vv b_1 ~.
\end{equation}
This is replaced by
\begin{equation}
\begin{aligned}
\vv x_0 \m A'_1&= (\vv x_0, 1)  \mbegin{cc}
\vv A_1 &  -\vv A_1\\
\vv b_1  & -\vv b_1
\mend  \\[0.1in]
&= \vv x _0\,[\vv A_1\,~ | - \vv A_1\,] + (\,\vv b_1, -\vv b_1) \\[0.1in]
&=
(~
\vv x_0\vv A_1 + \vv b_1 , -\vv x_0 \vv A_1 - \vv b_1
~)\\
&= (\vv x_0 \m A_1, - \vv x_0 \m A_1)
\end{aligned}
\end{equation}
In other words, the $m_0 \times m_1$ matrix $\vv A$ is replaced
by the $m_0 \times 2 m_1$ matrix $\vv A_1' =  [\,\vv A_1\,~ | - \vv A_1\,]$, and
$\vv b_1$ is replaced by the $2m_1$ dimensional vector $\vv b_1' = (\vv b_1, - \vv b_1)$.%
\footnote{
{\bf Important note: }
The affine transformation $\m A'$ is defined
\[
\mbegin{cc} \vv A_1 &  - \vv A_1 \\
\vv b_1 & - \vv b_1
\mend ~.
\]
This will output a vector of the form $(\vv x_0 \m A_1, - \vv x_0 \m A_1)$ with two parts
that are negatives of each other.
It defines the {\bf initial} form of the affine transform only.  During training,
all entries of the above matrix are free to vary (and will) independently, and its output will not maintain this symmetric form. 
This applies to all the affine transforms $\m A'_i$ that will be defined.
}

Now, when this is passed through an activation layer, represented by
the activation function $\sigma$ (ReLU), the result is
\begin{equation}
\begin{aligned}
\vv x_0 \m A_1'\sigma &= 
(~
\vv x_0 \m A_1  \sigma ~,  (-\vv x_0 \m A_1)\sigma ) = \vv x_1'   ~.
\end{aligned}
\end{equation}
This is the output of our modified affine layer with activation.
The point to note here is the identity $\vv v\sigma - (- \vv v)\sigma = \vv v$.  This gives
\begin{equation}
\label{eq:complementarity}
(\vv x_0 \m A_1) \sigma -  (-\vv x_0 \m A_1)\sigma = \vv x_0 \m A_1 ~.
\end{equation}
Thus, by subtracting the two halves of $\vv x_0 \m A_1'\sigma$, one arrives back at
the simple affine transform $\vv x_0 \m A_1$.

We simplify the notation as follows.  Let  $\vv x_i^+$ and $\vv x_i^-$ be defined as
\begin{subequations}
\begin{align}
\vv x_i^+ &= \vv x_i\sigma \\
\vv x_i^-  &= (- \vv x_i)\sigma
\end{align}
\end{subequations}
which are the positive and negative parts of $\vv x_i$ respectively, satisfying
$\vv x_i = \vv x_i^+ - \vv x_i^-$.
Then, we see that%
\begin{equation}
\label{eq:induction0}
 \vv x'_1 =  (\vv x_1^+, \vv x_1^-)  ~.
\end{equation}

Now, the first thing to do at the beginning of the next layer is to subtract the two parts
of the previous output, illustrated by
\begin{equation}
\label{eq:intermediate1}
\vv x_1' \mbegin{r}\m I \\ - \m I \mend 
= (\vv x_1^+, \vv x_1^-)  \mbegin{r}\m I \\ - \m I \mend = \vv x_1 ~.
\end{equation}
This is followed by the same trick of separating the positive and negative
parts as before.
The affine part (without activation) of the second layer is then
\begin{align}
\begin{split}
\label{eq:recursion}
\vv x'_1 \m A_2' &= \vv x_1'\mbegin{r}\m I \\ - \m I \mend 
[\vv A_2 ~|~ -\vv A_2] + (\vv b_2, - \vv b_2) \\
&= 
\vv x_1' 
\mbegin{rr}\vv A_2 & - \vv A_2\\ - \vv A_2 & \vv A_2 \mend + 
(\vv b_2, - \vv b_2) \\
&= \vv x_1' \vv A_2' + \vv b_2' \ ,
\end{split}
\end{align}
where $\vv A_2'$, $\vv b'_2$ and $\m A'_2$ are defined by this equation.
Noting \eq{intermediate1}, we arrive at
\begin{equation}
\vv x_1' \m A_2' = (\vv x_1 \m A_2, - \vv x_1 \m A_2) ~.
\end{equation}
and so
\begin{equation}
\label{eq:induction1}
 \vv x_1' \m A_2' \sigma = \vv x_2' =(\vv x_1 \m A_2\sigma, (-\vv x_1 \m A_2)\sigma) = (\vv x_2^+, \vv x_2^-) ~.
\end{equation}
Combining  this with \eq{induction0} gives
\begin{equation}
\vv x_0\, \m A'_1\sigma \, \m A'_2\sigma =  (\vv x_2^+, \vv x_2^-) ~.
\end{equation}
Continuing to define layers in this way, according to \eq{recursion}, for
layers up to layer $k-1$, gives that
\begin{equation}
\vv x_0\, \m A'_1\sigma \, \m A'_2 \sigma\ldots \m A'_{k-1} \sigma= (\vv x_{k-1}^+, \vv x_{k-1}^-)  = \vv x_{k-1}' ~.
\end{equation}
Finally, we define the last layer $k$ by

\begin{align}
\begin{split}
\label{eq:recursion}
\vv x'_{k-1} \m A_k' &= (\vv x_{k-1}^+, \vv x_{k-1}^-) \mbegin{r}\vv A_k\\ - \vv A_k\mend 
+ \vv b_k \\
&= (\vv x_{k-1}^+ - \vv x_{k-1}^- )\vv A_k  + \vv b_k\\
&= \vv x_{k-1} \vv A_k  + \vv b_k\\
&= \vv x_{k-1} \m A_k \\
&=\vv x_k ~.
\end{split}
\end{align}

Putting those defined $\m A_i$ all together gives
\begin{equation}
\vv x_0 \,\m A_1' \sigma \,\m A_2 \sigma \ldots \m A_k' = \vv x_k = \vv x_0\ ,
\end{equation}
where the final step is because the sequence of affine transforms
$\m A_1 \ldots \m A_k$ is chosen to be the identity map,
so $\vv x_k = \vv x_0 \m A_1 \ldots \m A_k = \vv x_0$.

\section{Exploration of Layer Initialization Effects} \label{sec:loss_curves}

\begin{figure}[t]
\begin{center}
\includegraphics[width=0.46\textwidth]{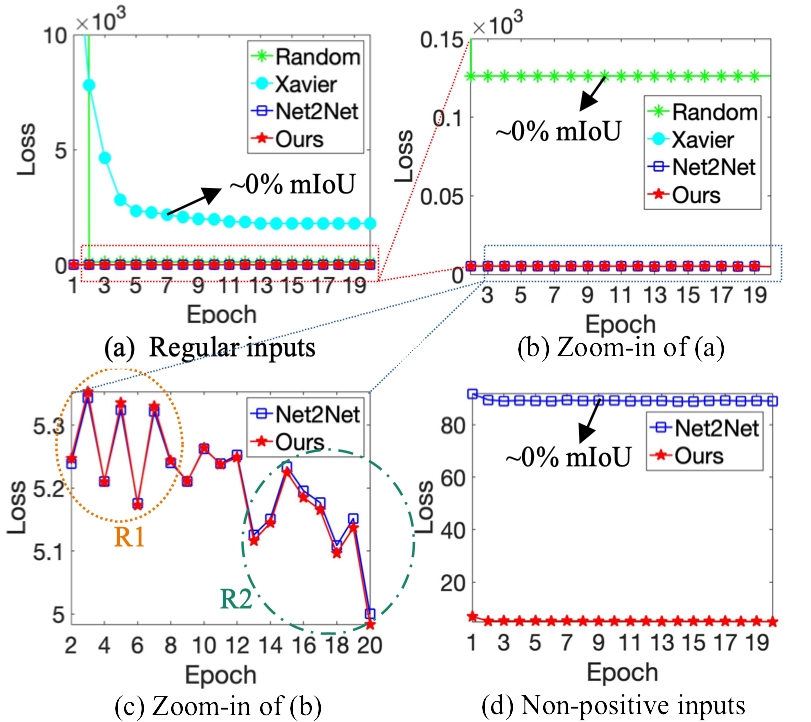}
\caption{\small \em Training loss with different layer initialization methods.
This is an extended ablation study to Tables~\myred{2} and \myred{4} in the main
paper.
``Regular": values containing positive, negative, and zero values.
Note that in our semantic segmentation task, a low loss is determined by a high (mostly positive) logit from the NN along the label dimension.
The joint learning is for 21-label semantic segmentation using pretrained DeepLabV3+ \cite{deeplabv3+} and
superpixel with FCN \cite{sp_fcn} networks with add-on 3 Fully-Connected (FC) layers.
\textbf{Here, in (a) and (b), random and Xavier initialization on these add-on FC layers lead to a high loss, and thus, decreasing the mIoU from $\sim$80\% to $\sim$0\%.} It is obvious that they cannot work in our case, since they are unable to recover the pretrained results as shown in Table~\myred{2} in the main paper. On the other hand, Xavier initialization may have a worse local minima than random initialization due to the interrupted output values.
\textbf{In contrast, for regular data containing positive, negative, and zero values, both Net2Net and our transparent initialization have similar good effectiveness.} Because, as mentioned before, negative logit values (hardly to be the highest) may not have significant effects on the joint learning as in our case positive logits always have high softmax values. So, the negative values of the input data can even be zero-out by ReLU or Net2Net initialization. \textbf{For non-positive inputs in (d), however, Net2Net is unable to recover negative values, leading to $\sim$0\% mIoU while ours has the same low loss as it is in (c).}
Additionally, although Net2Net and ours in (c) have similar losses, both achieving $\sim$83.3\% mIoU, we note that the loss of ours starts to be less than Net2Net, shown in $R2$. In $R1$, ours has a high loss due to the effects of dense gradients that change network parameters more dramatically than Net2Net. This is expected as the transparent initialization should have a strong learning ability than Net2Net.
\textbf{Overall, ours outperforms random and Xavier initialization, both regular and non-positive data, and Net2Net for non-positive data. For regular data (depends on task), ours tends to have a smaller loss than Net2Net due to its strong learning ability with dense gradients.}
More details are in Sec.~\ref{sec:loss_curves}.
}
\label{fig:loss_curve}
\end{center}
\end{figure}

In Table~\myred{2} in the main paper, we compared the effectiveness of
our transparent initialization with others, that is random,
Xavier \cite{xavier}, and Net2Net \cite{net2net}.
Transparent initialization can 100\% recover the input data from the layers
output with a high initialization rate and a small around-off error
shown in Table~\myred{3} and supports
non-square filters, by saying filters we mean $1^2$ kernel-size
convolutional layers or fully-connected layers.
In contrast, random and Xavier initialization are ineffective to recover
the input data from the output while Net2Net is only effective for
non-negative data and is infeasible for non-square filters.
Despite these quantitative experiments to analyse the attributes of
transparent initialization, we directly compare their effects on joint
learning pretrained networks of semantic segmentation and superpixels
in addition to the ablation study in the main paper.

In Fig.~\ref{fig:loss_curve}, for the task of semantic segmentation, the cross-entropy loss function highly relies on the maximum values of logits, the network outputs, along the label dimension.
In Fig.~\myred{9(a)}-\myred{9(c)}, input data of the add-on FC layers contains positive and non-positive values while in Fig.~\myred{9(d)} input data is non-positive by subtracting the maximum value along the label dimension.
This is to explore the data recovering ability for different numerical space (that is non-negative and non-positive in our case).

The final mIoUs via random and Xavier initialization are nearly 0\% as they totally interrupt the learned parameters, leading to a high loss. To recover \textbf{non-negative} data, in Fig.~\myred{9(c)}, Net2Net and ours have similar loss decreasing tendency. However, as the epoch increases, the loss by ours becomes lower than Net2Net because the gradients by transparent initialization are dense for parameter updates compared with the sparse ones by Net2Net. Furthermore, in Fig.~\myred{9(d)}, Net2Net is unable to recover \textbf{negative} values, as shown in Table~\myred{2} in the main paper, resulting in a high loss.
In Fig.~\myred{9(c)}, the mIoU by Net2Net and ours are similar, both nearly 83.3\%.
Corresponding to Fig.~\myred{9(d)}, however, the mIoU by Net2Net is nearly 0\% while ours is still nearly 83.3\% since it is invariant to numerical space.

\section{Activation Functions}
\label{sec:activations}
Again, note that the notation of activation function $\sigma(x)$ has the same meaning as
$x \sigma$ for a simplicity of sequence layers.
For the proposed transparent initialization, any non-linear
activation functions satisfying Eq.~\eqref{eq:recover} are feasible
to recover the input data from its output.
\begin{equation}
\sigma(x) - \sigma(-x) = cx\ ,
\label{eq:recover}
\end{equation}
where $c$ is a non-zero constant.
This can be expressed by $x D \sigma S$ with $D$ (duplicate) and
$S$ (subtract) defined in the main paper.
In Fig.~\ref{fig:act}, we give 4 examples with corresponding definitions
as follows:
\begin{subequations}
\begin{align}
\text{ReLU:}&\quad y(x) =
    \left\{
    \begin{array}{ll}
    x  & \mbox{if $x \ge 0$}\ ,\\
    0  & \mbox{otherwise}\ ,
    \end{array}
    \right. \\
\text{LeakyReLU:}&\quad y(x) =
    \left\{
    \begin{array}{ll}
    x  & \mbox{if $x \ge 0$}\ ,\\
    \delta x  & \mbox{otherwise}\ ,
    \end{array}
    \right. \\
\text{SoftReLU:}&\quad y(x)
= \log \left( 1 + e^{x} \right)\ , \\
\text{LogSigmoid:}&\quad y(x)
= \log \left( \frac{1}{1 + e^{-x}} \right)\ .
\label{eq:act_define}
\end{align}
\end{subequations}

\begin{figure}[t]
\begin{center}
\begin{subfigure}[b]{0.2\textwidth}
\includegraphics[width=\textwidth]{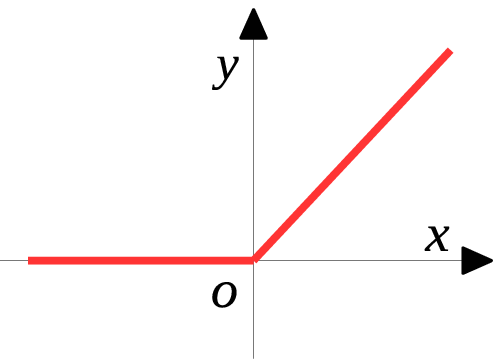}
\vspace{-1\baselineskip}
\caption{\small ReLU}
\label{fig:relu}
\end{subfigure}
\hspace{2mm}
\begin{subfigure}[b]{0.2\textwidth}
\includegraphics[width=\textwidth]{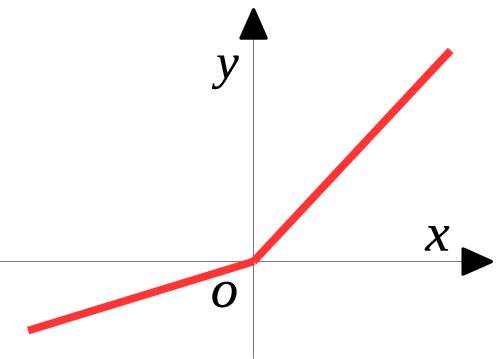}
\vspace{-1\baselineskip}
\caption{\small LeakyReLU}
\label{fig:leakyrelu}
\end{subfigure}
\hspace{2mm}
\begin{subfigure}[b]{0.2\textwidth}
\includegraphics[width=\textwidth]{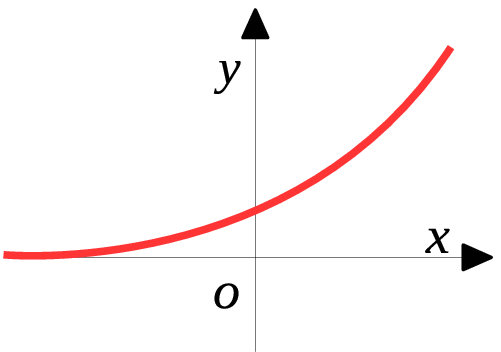}
\vspace{-1\baselineskip}
\caption{\small SoftReLU}
\label{fig:softrelu}
\end{subfigure}
\hspace{2mm}
\begin{subfigure}[b]{0.2\textwidth}
\includegraphics[width=\textwidth]{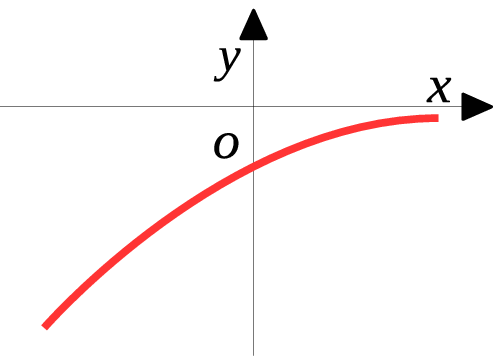}
\vspace{-1\baselineskip}
\caption{\small LogSigmoid}
\label{fig:logsigmoid}
\end{subfigure}
\end{center}
\vspace{-\baselineskip}
\caption{\small\em Examples of non-linear active functions for transparent
initialization.}
\label{fig:act}
\end{figure}

It is easy to verify Eq.~\eqref{eq:recover} for those activation functions.
Given the activation function as LogSigmoid by $\sigma(x)=\log(1/(1+e^{-x}))$, for instance, it follows
\begin{equation}
\begin{aligned}
&\log \left( \frac{1}{1 + e^{-x}} \right) - \log \left( \frac{1}{1 + e^{x}} \right)\\
=&\log \left( \frac{1+e^x}{1 + e^{-x}} \right)
=\log(e^x)
=x\ .
\end{aligned}
\end{equation}

Meanwhile, with $c$ given in Eq.~\eqref{eq:recover}, $x D \sigma S$
should be $x D \sigma S / c$, where
$c=1/\delta$ is for LeakyReLU and $c=1$ for the others.
To be more general, this property holds for any functions of the
form of $f(x) = cx + s(x)$, where $s(x)$ is an even function.

For other activations $\sigma$ such as the sigmoid function, or
arctangent, or hyperbolic tangent, it does not work directly.
The trick is to observe that for the ReLU function,  $(-x)\sigma = x\sigma - x$.
This allows us to rewrite the sequence of operations
\[
x \xmapsto {\sigma} (x\sigma, (-x)\sigma) = (x_1, x_2) \mapsto x_1 - x_2 = x ~,
\]
where $\sigma$ is the ReLU function, as
\[
x \xmapsto{\sigma} (x \sigma, x \sigma - x) = (x_1, x_2) \mapsto x_1 - x_2 = x ~.
\]
But now this also holds for any activation $\sigma$,
ReLU or not.  This requires a slight change to the architecture of the 
transparent affine layer.  Instead of outputting $(x\sigma, (-x)\sigma)$,
it must output $(x\sigma, x\sigma - x)$.  This is shown in Fig.~\ref{fig:different-activation}.
\begin{figure}[t]
\centering
\begin{subfigure}[b]{0.26\textwidth}
\includegraphics[width=\textwidth]{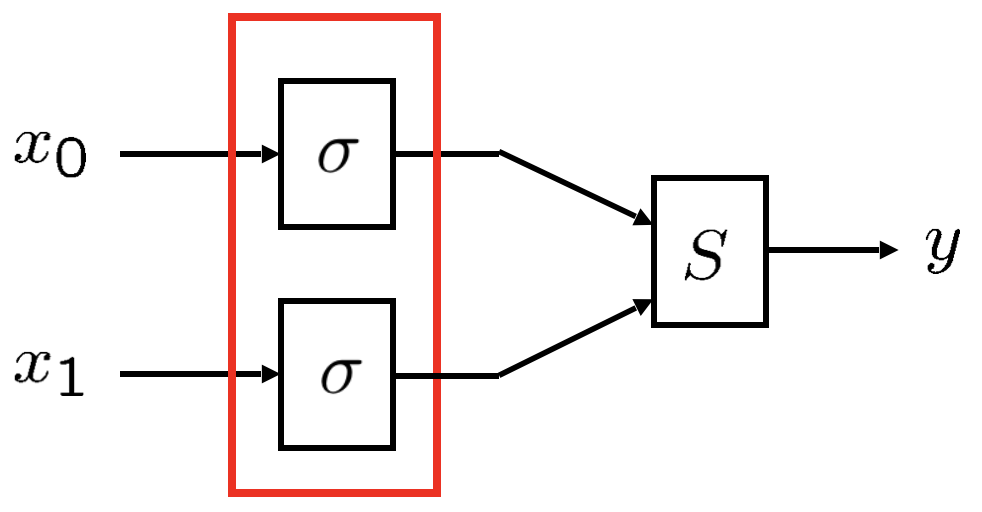}
\subcaption{For $y = (x_0\sigma, x_1\sigma)\, S$}
\vspace{3mm}
\end{subfigure}
\begin{subfigure}[b]{0.32\textwidth}
\includegraphics[width=\textwidth]{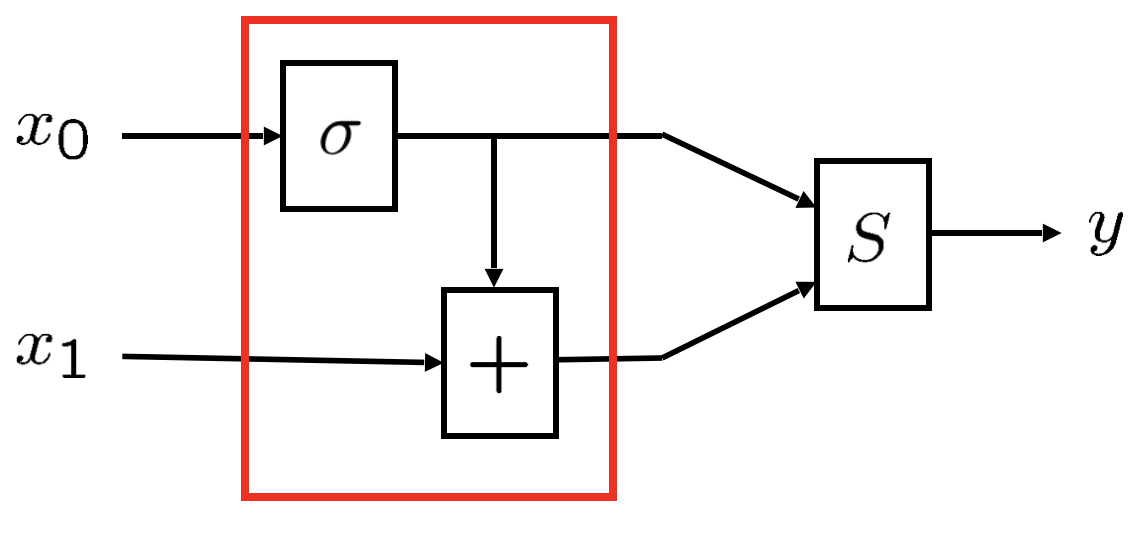}
\subcaption{For $y = (x_0 \sigma, x_0\sigma + x_1) S$}
\end{subfigure}
\caption{\small\em Two different forms of layer activation (the part in the red box)
top be used as see-through activations for transparent layers.
Here, $\sigma$
represents any function, for instance, a commonly used non-linear function
such as sigmoid, ReLU or hyperbolic tangent, and the block marked $S$ carries
out some operation on the
two inputs.  Thus, (a) implements $y = (x_0\sigma, x_1\sigma)\, S$ and
(b) represents $y = (x_0 \sigma, x_0\sigma + x_1) S$.
At initialization, 
 $x_0 = -x_1 = x$ and $(a, b)S = a-b$.  In this case, the output is $y = x$
 in both case.  During the training of the network, $x_0$ and $x_1$ will diverge,
and $S$ will also learn to carry out a different operation, so networks 
will behave differently.   However, using (b)
to implement the affine layer activation will provide a transparent initialization,
whatever function $\sigma$ is chosen.
}
\label{fig:different-activation}
\end{figure}

\section{Extension to Convolutional Layers}
\label{sec:convolutional}
A common practice in experimenting with modifications to neural network architectures
is to add additional convolutional layers. If a base network is already trained to
achieve minimal loss, then the addition of extra convolutional layers can perturb
the loss, and require the network to retrain again to achieve a low loss.
The loss after addition of the new layers may not be as low as the loss of the base
network. However, by initializing the extra add-on layers to implement an identity
transformation will ensure that the loss achieved (on the training set) by the modified
network cannot be worse than that of the base network.

This section gives some ideas on the addition of transparent convolutional layers
with pseudo-random initialization that could be used for this purpose.  We have
not explored this topic in any great depth, leaving it rather to be the subject of
future work, and is somewhat speculative at present.

A convolutional layer as usually implemented is an example of an affine
layer, since without any non-linear activation it implements a linear transformation
on the input and then adds bias.  However, unlike fully-connected layers, which
are essentially a matrix multiplication, for which an inverse (or right-inverse) can be
easily found, convolution is not conveniently represented by matrix multiplication
and is less easily inverted.

It is clear that convolution, being linear, can be represented as matrix multiplication 
on a vectorized form of the input, but this will be a sparse matrix multiplication, 
and not easily inverted, at least by convolutional layers.
By analogy with the method implemented for fully-connected layers, we require
that a final convolution will invert the effect of a sequence of previous
pseudo-random convolutions.  

Not all convolutions are exactly invertible.  In fact (ignoring edge effects) 
a convolution will be invertible by another convolution if and only if 
its Fourier transform is everywhere non-zero.  However, with this caveat, it
is possible to find inverse convolutions.  

Rather than carrying out a string of convolutions, followed by a single convolution
to undo the previous ones, a better strategy may be to add convolutions in
pairs such as a high-pass and low-pass filter that cancel each other out.
Exploring this topic will be the subject of further work.
\else

\begin{abstract}
Although deep learning greatly improves the performance
of semantic segmentation, its success mainly lies in object central areas
without accurate edges.
As superpixels are a popular and effective auxiliary to preserve object
edges, in this paper, we jointly learn semantic segmentation with trainable
superpixels.
We achieve it with fully-connected layers with Transparent
Initialization (TI) and efficient logit consistency using a sparse encoder.
The proposed TI preserves
the effects
of learned parameters of pretrained networks.
This avoids a significant increase of the loss of pretrained networks,
which otherwise may be caused by inappropriate parameter
initialization of the additional layers.
Meanwhile, consistent pixel labels in each superpixel
are guaranteed by logit consistency.
The sparse encoder with sparse matrix operations substantially reduces both the memory requirement and the computational complexity.
We demonstrated the superiority of TI over
other parameter initialization methods and tested its numerical stability.
The effectiveness of our proposal was validated on PASCAL VOC 2012, ADE20K, and PASCAL Context
showing enhanced semantic segmentation edges.
With quantitative evaluations on segmentation edges using
performance ratio and F-measure, our method outperforms the 
state-of-the-art.
\end{abstract}

\section{Introduction}

Semantic segmentation is an essential task in computer
vision which requires mapping image pixels of interesting objects to predefined semantic labels.
Applications include autonomous driving 
\cite{auto_driving}, object identification \cite{pcontext,detection},
image editing \cite{editing}, and scene
analysis \cite{scene_understanding}.
Recent developments of semantic segmentation are greatly promoted by deep
learning on several large-scale datasets \cite{berkeley,coco}, resulting in effective networks \cite{fcn,psp,crfasrnn,deeplabv1,deeplabv2,deeplabv3+,encnet,resnest}.
Semantic segmentation obtained by these methods, however, is not substantially
aligned with object edges.


This problem can be alleviated by using qualified edge-preserving 
methods \cite{densecrf,crfasrnn,deeplabv1}
or independently learning edges for semantic segmentation
\cite{ssn,sp_bi,deeplabv2}.
Nevertheless, those edges are usually incomplete and oversegmented.
To solve this problem, denseCRF \cite{densecrf} based methods
aggregate object features over a large and dense range using bilateral
filters with high-dimensional lattice computations.
This, however, could be more efficient and desirable to learn on locally oversegmented
areas, such as superpixels that aggregate similar pixels into a higher-order
clique \cite{slic,lsc}.

Existing superpixel segmentation approaches are mainly categorized into
traditional methods, such as SLIC \cite{slic}, LSC \cite{lsc},
Crisp \cite{crisp}, BASS \cite{bass}, and those
using neural networks, such as SSN \cite{ssn},
affinity loss \cite{seal}, and superpixels by FCN \cite{sp_fcn}.
Although those traditional methods have qualified performance on superpixel
segmentation, they cannot be easily embedded into neural networks for
end-to-end learning due to nondifferentiability or large
computational complexity.
In contrast, \cite{ssn} provides an end-to-end learning for
superpixel semantic segmentation, but the pixel labels in each superpixel are not always consistent.
Similarly, \cite{sp_fcn} uses a fully convolutional network to learn
superpixels for stereo matching, which is the
current state-of-the-art in superpixel learning.
However, the problem of inconsistent pixel labels in each superpixel will also occur when \cite{sp_fcn} is applied
to semantic segmentation.

To deal with the problems of edge loss and inconsistent superpixel labelling,
we jointly train
these networks with additional fully-connected
layers using transparent initialization and logit consistency,
resulting in enhanced object edges in Fig.~\ref{fig:demo}.
Specifically, transparent initialization warm starts the training
by recovering the pretrained network output from its input at initialization, followed by a
gradual improvement in learning superpixels.
Simultaneously, logit consistency with sparse encoder enables 
efficient logit averaging to guarantee consistent pixel labels in each superpixel.
Furthermore, we used the popular performance ratio \cite{pr_metric} and F-measure \cite{f_measure_1,f_measure_2} to evaluate
the quality of semantic segmentation edges.
Our code will be available upon publication.
The main contributions of this work are:
\vspace{-0.3\baselineskip}
\begin{itemize}
\item We jointly learn the state-of-the-art semantic segmentation network and
superpixel network to enhance the labelling performance with
sharp object edges.
The improvement is vivid as demonstrated by evaluating on PASCAL VOC 2012,
ADE20K and PASCAL Context.

\vspace{-0.3\baselineskip}
\item Transparent initialization is proposed  for learning fused features
while adding fully-connected layers to pretrained networks.
This helps to join and fine-tune state-of-the-art networks without interrupting
the effect of pretrained network parameters.
The transparent initialization may also be adapted to fine-tune
multiple and deeper pretrained networks for other tasks.

\vspace{-0.3\baselineskip}
\item Logit consistency, implemented by a sparse encoder with sparse matrix
operations, ensures consistent semantics for all
pixels in each superpixel.
This makes it feasible and efficient to index pixels
by superpixels, because of greatly reduced computational complexity.
\end{itemize}

\section{Related Work}

\paragraph{Semantic Segmentation.}
Semantic segmentation can be traced back to early
techniques \cite{semantic_survey} based on classifiers, such as
random decision forests \cite{random_forest}, SVMs \cite{svm_seg}, and
graphic models with MRFs and CRFs \cite{jordan_book,densecrf,ajanthan2017efficient}.
In contrast, modern state-of-the-art methods rely on
advanced exploitation of deep CNN classifiers, such as ResNet \cite{resnet},
DenseNet \cite{densenet}, and VGG16 \cite{vgg16}.
Fully Convolutional Network (FCN) \cite{fcn} and related methods are typical
architectures that use rich image features from classifiers
usually pretrained on ImageNet \cite{imagenet}.
SegNet \cite{segnet} uses a U-Net structure for an encoder-decoder module
to compensate for low resolution by using multiple upsampled feature maps.
In addition, several multiscale contextual fusion methods
\cite{deeplabv2,refinenet,deeplabv3+} have been proposed to
aggregate pyramid feature maps for fine-grained segmentation.
Typically, DeepLabV3+ \cite{deeplabv3+} combines spatial pyramid pooling
and encoder-decoder modules to refine the segmentation along with 
object boundaries.
Recently, attention-based networks \cite{emanet,psanet,a2net,danet}
improve object labelling confidence by aggregating features of a
single pixel with those from other positions.
Zhang \textit{et al.} \cite{resnest} introduce a
split-attention block in ResNet (ResNeSt), which enables multiscale
scores with softmax for attention across feature-map groups.
This achieves the new state-of-the-art
performance in
semantic segmentation and image classification.
Due to the limited capability
of network architectures, however,
a huge improvement on mean Intersection over Union (mIoU)
is hard to achieve; see the
minor improvement in \cite{region_seg}.

\vspace{-2.5ex}
\paragraph{Superpixel Segmentation.}
Superpixel segmentation has been well studied for years with a
comprehensive survey in \cite{sp_survey}.
In contrast to classical methods that initialize superpixel regions with seeds
and cluster pixel sets using distance measurement \cite{distance},
boundary pixel exchange \cite{exchange}, etc., the widely-used
SLIC based methods \cite{slic,manifold_slic,lsc,snic} employ
(weighted) K-means clustering
on pixel feature vectors to group neighbouring pixels.
In deep learning, SSN \cite{ssn} first proposes an end-to-end learning
framework for superpixels with differentiable SLIC for semantic segmentation
and optical flow.
By comparison, \cite{sp_fcn} replaces the soft K-means manner in SSN with a
simple fully convolutional network and applies it to stereo matching with
downsampling and upsampling modules.
While SSN results in superpixel-level semantic segmentation, the pixel
labelling is not aligned with the superpixels.
For instance, inconsistent labels exist in each superpixel,
which reduces the effects of superpixels on semantic assignments.

\vspace{-2.5ex}
\paragraph{Superpixel Semantic Segmentation.} Some works use
superpixels to optimize graph relations \cite{sp_steve,sp_graph} or
downsample images as a pooling alternative to max or average pooling
\cite{sp_bi,sp_control,sp_efficient}.
These methods usually use fixed superpixels obtained by
traditional methods mentioned above (``Superpixel Segmentation'')
and lose the exact alignment of segmentation edges with superpixel
contours after upsampling.
This, however, can be easily achieved by our logit consistency module.
Moreover, fixed superpixels are unsuitable for end-to-end learning since
traditional methods, such as SLIC \cite{slic}, are computationally expensive
due to CPU execution and inflexible for fine-tuning superpixels around
object edges, especially for small objects.
Instead, superpixels learned by CNN alleviate this problem.

\section{Methodology}

The flowchart of our method is shown in Fig.~\ref{fig:flowchart}.
Subsequently, we discuss each of these modules in detail.

\begin{figure}[t]
\centering
\includegraphics[width=0.47\textwidth]{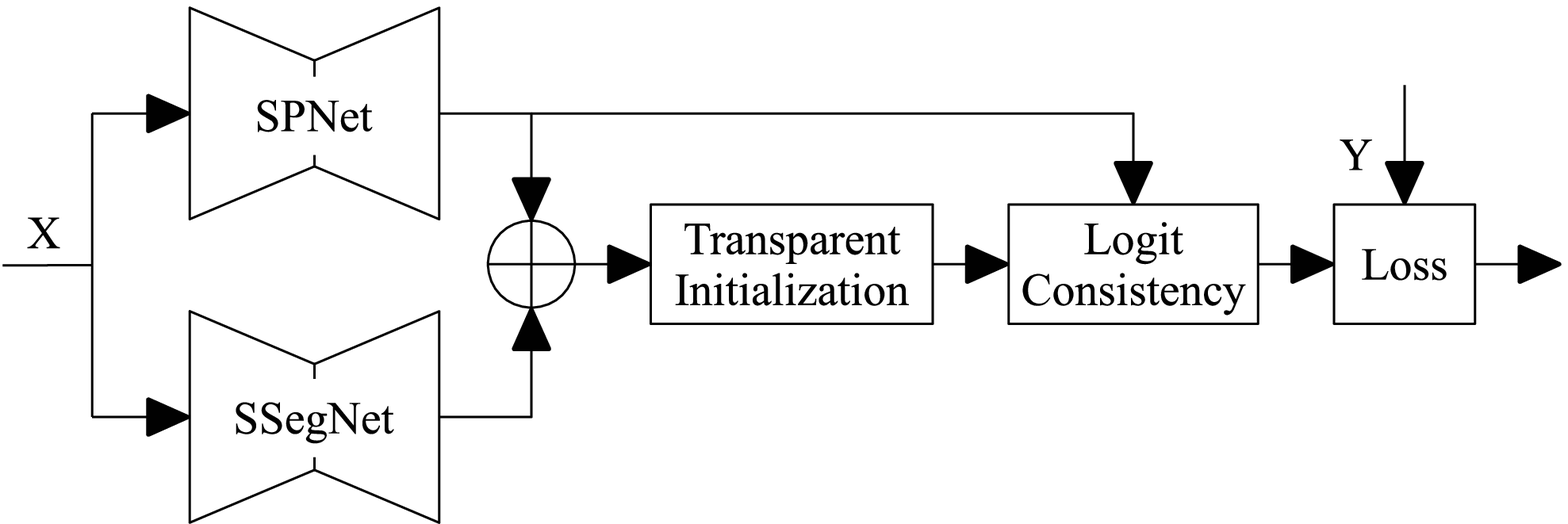}
\vspace{-2mm}
\caption{\em\small Flowchart.
\textbf{X:} input image,
\textbf{Y:} semantic segmentation ground truth,
\textbf{SPNet:} superpixel network,
\textbf{SSegNet:} semantic segmentation network,
\textbf{$\bigoplus$:} concatenatation,
\textbf{Transparent Initialization:} identity mapping by fully-connected layers,
\textbf{Logit Consistency:} consistent pixel labels in each superpixel.
}
\label{fig:flowchart}
\vspace{-3mm}
\end{figure}

\subsection{Network Architectures}
\label{sec:networks}
We use the state-of-the-art DeepLabV3+ \cite{deeplabv3+} and
ResNeSt \cite{resnest} for semantic segmentation and the state-of-the-art superpixels
with FCN \cite{sp_fcn} for superpixel contours.

DeepLabV3+ achieves high and robust performance with atrous spatial
pyramid pooling for multiscale feature maps and
encoder-decoder modules for deep features with
different output strides.
It is widely used as a network backbone for semantic segmentation due to these
well-studied modules evaluated by empirical experiments.
ResNeSt \cite{resnest} replaces ResNet with multiscale scores using softmax-based feature map attention and achieves the new state-of-the-art.
The loss function for semantic segmentation is the standard
cross-entropy loss of predicted logits and ground truth labelling.
Readers can refer \cite{deeplabv3+,resnest} for more details.

Superpixel with FCN \cite{sp_fcn} is the current state-of-the-art superpixel network.
The core idea is to construct a distance-based loss function with
aggregations of neighbouring pixel and superpixel properties and
locations.
This is similar to the SLIC method \cite{slic}, where the property
vectors can be CIELAB colors or one-hot semantic encoding.
The loss function for superpixel network with FCN \cite{sp_fcn}
is in Eq.~\eqref{eq:sp_fcn}.

Given an image with $N_p$ pixels and $N_s$ superpixels,
we denote the subsets of pixels as
$\mathcal{P}=\{\mathcal{P}_0, ..., \mathcal{P}_{N_s-1}\}$, where
$\mathcal{P}_i$ is a subset of pixels belonging to superpixel $i$.
With pixel property $\mathbf{f} \in \mathbb{R}^{N_p \times K}$
($K$ features for each pixel)
and probability map $\mathbf{q} \in \mathbb{R}^{N_s \times N_p}$,
the loss function is
\begin{equation}
\begin{aligned}
\label{eq:sp_fcn}
\mathcal{L}(\mathbf{f},\mathbf{q})
= \sum_{p \in \mathcal{P}} E \left( \mathbf{f}(p), \mathbf{f}'(p) \right)
+ \frac{m}{D} \vert \vert \mathbf{c}(p) - \mathbf{c}'(p) \vert \vert_{2}
\end{aligned}
\end{equation}
with
\begin{subequations}
\begin{align}
\mathbf{u}_{s}
= \frac{\sum_{p \in \mathcal{P}_s} \mathbf{f}(p) q_{s}(p)}
{\sum_{p \in \mathcal{P}_s} q_{s}(p)}&,
\quad \mathbf{l}_{s} = \frac{\sum_{p \in \mathcal{P}_s} \mathbf{c}(p)
q_{s}(p)} {\sum_{p \in \mathcal{P}_s} q_{s}(p)}, \\
\mathbf{f}'(p) = \sum_{s \in \mathcal{N}_p} \mathbf{u}_s
q_s(p)&,
\quad \mathbf{c}'(p)
= \sum_{s \in \mathcal{N}_p} \mathbf{l}_s q_s(p),
\end{align}
\end{subequations}
where $m$ is a weight balancing the effects of property and coordinates
on loss, $D$ is a superpixel sampling interval in proportion to superpixel
size, $\mathbf{c}(i)=[x_i, y_i]^T$ for all $i \in \{1, ..., N_p\}$ are
pixel coordinates, $E(\cdot, \cdot)$ is a distance measure function
involving $l_2$ norm or cross-entropy, $q_s(p)\in \mathbf{q}$ is the
probability of pixel $p$ belonging to superpixel $s$, and $\mathcal{N}_p$
is a set of superpixels surrounding pixel $p$.

Here, $\mathbf{u}_s$ and $\mathbf{l}_s$ are superpixel-level property
vector and central coordinates aggregated from involved pixels
respectively, and $\mathbf{f}'$ and $\mathbf{c}'$ are pixel-level property
and coordinates aggregated from surrounding superpixels.
This updates between pixels and superpixels until the loss
converges.

\begin{figure}[t]
\begin{center}
\includegraphics[width=0.48\textwidth]{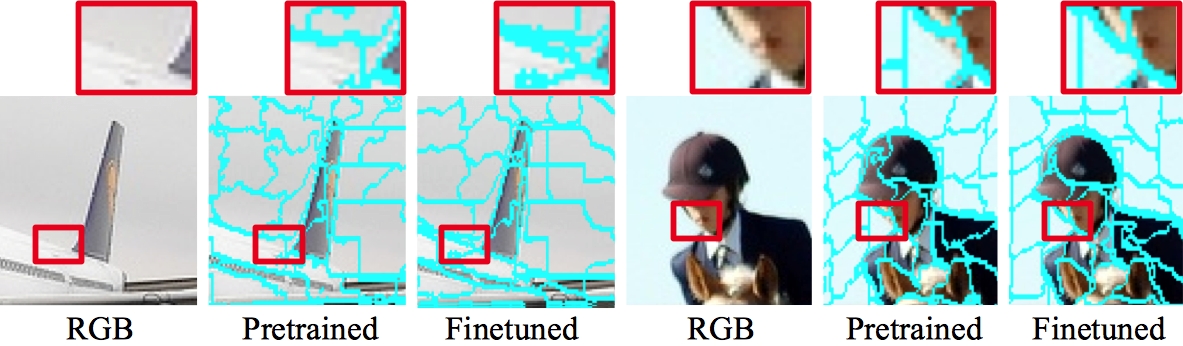}
\vspace{-1.5\baselineskip}
\caption{\em\small Pretrained superpixles on BSDS500 \cite{sp_fcn}
\textit{vs.} fine-tuned superpixels on PASCAL VOC 2012 using our joint learning.
Fine-tuned superpixels recover accurate object edges to alleviate the domain
gap between datasets.
Best view by zoom-in.}
\label{fig:fixed_vs_learned}
\end{center}
\vspace{-8mm}
\end{figure}

\subsection{Learning with Transparent Initialization}

Although some semantic segmentation datasets, such as PASCAL VOC \cite{voc2012} and
Berkeley benchmark \cite{berkeley}, have no accurate edges for supervised edge learning,
it can be compensated for by a superpixel
network on edge-specified datasets, such as SBDS500 \cite{sbds500}.
Due to the domain gap of datasets, however, 
pretrained superpixel models are more desirable than learning
from scratch for fast loss convergence.
Superpixel maps pretrained on BSDS500, however, are not always
suitable for semantic segmentation datasets.
Hence, fine-tuning by joint learning is necessary to improve the quality of
superpixel contours, compared with using the pretrained network,
as shown in Fig.~\ref{fig:fixed_vs_learned}.

Therefore, in our proposal, linear layers or convolutional layers with
$1 \times 1$ kernels
are added to fuse the outputs of superpixel and semantic segmentation
networks.
Either or both can be pretrained.
Importantly, selecting an appropriate initialization on these
additional layers is important to avoid overriding the effect of
learned network parameters.
It is straightforward to cast the layer operations as an
identity mapping between input and output at the early training stage.
Net2Net \cite{net2net} achieves this by using identity matrices to
initialize linear layers.
This method, however, is inefficient in learning since the identity
matrices will result in highly sparse gradients.
In addition, it cannot handle activation functions with negative values.

In contrast, we introduce {\em transparent initialization} with non-zero
values for dense gradients,
while identically mapping the layer input to output and preserving the
effect of the learned parameters of the pretrained networks.

\vspace{-1ex}
\paragraph{Affine Layers.}
A linear layer {without activation} (such as a convolution or
fully-connected layer) can be written as $\mathbf{y}=
\mathbf{x} \mathbf{A} + \mathbf{b}$ with layer weights
$\mathbf{A}$, bias $\mathbf{b}$, and input $\mathbf{x}$,
by matrix multiplication with
$\tilde{\mathbf{y}}=[\mathbf{y}, \mathbf{1}]$:
\begin{equation}
\label{eq:conv}
\begin{aligned}
\tilde{\mathbf{y}}
= \tilde{\mathbf{x}} M
= \left[ \mathbf{x}, \mathbf{1} \right]
    \begin{bmatrix}
    \mathbf{A} & \mathbf{0} \\
    \mathbf{b} & \mathbf{1}
    \end{bmatrix}.
\end{aligned}
\end{equation}
For simplicity the mapping \eq{conv} will be denoted as $\vv y = \vv x\, \m A$
where $\m A$ denotes the affine transformation, and we place the 
functions on the right.
Note the difference between $\m A$, an affine transform
and $\mathbf A$, a matrix.
If an activation function $\sigma$ is included, then the mapping
is $\vv x \mapsto \vv x \,\m A \,\sigma$.

The right-inverse of an affine transformation  \eq{conv} is 
represented by the matrix
\begin{equation}
\label{eq:right_inverse}
\begin{aligned}
M^{R}
= \begin{bmatrix}
    \mathbf{A}^{R} & \mathbf{0} \\
    -\mathbf{b}\mathbf{A}^R & \mathbf{1}
    \end{bmatrix},
\end{aligned}
\end{equation}
satisfying $M M^R = I$, the identity map.
Here, $\textbf A^R$ is the right-inverse of matrix $\mathbf A$, which exists
if $\textbf A$ has dimension $m\times n$ and rank $m$,
in which case
$\textbf A^R = \textbf A\tr (\textbf A \textbf A\tr)^{-1}$.


\vspace{-1ex}
\paragraph{Transparent Initialization.}
The idea behind transparent layer initialization is to construct
a sequence of $k\ge 2$ affine layers denoted
$\m A_1, \m A_2, \ldots, \m A_k$ such that $\m A_k$ is
a right-inverse of the product $\m A_1 \ldots \m A_{k-1}$, 
satisfying $\m A_1 \m A_2 \ldots \m A_k = \m I$, the
identity transformation. 

A necessary condition for this right inverse to exist is that $\m A_1 \m A_2 \ldots \m A_{k-1}$ should
be of full rank. 
\SKIP{A necessary condition for this to happen is as follows.
Let $\m A_i$ represent an affine transformation $\m A_i : \R^{m_{i-1}} \rightarrow \R^{m_i}$.
The first and last spaces in this sequence
are $\R^{m_0}$ and $\R^{m_k}$,
so $m_0 = m$ and $m_k = n$. 
Then $\m A_1 \ldots \m A_{k-1}$ has a right inverse
only if $m_i \ge m_0$ for all $i=1, \ldots, k-1$.  In other words, a necessary condition
for the right inverse to exist is that the dimensions of all the intermediate
spaces $\R^{m_1}, \ldots ,\R^{m_{k-1}}$ should be at least equal to $m_0$.
It is not hard to see that generically, this is also a sufficient condition,
meaning that it is true for almost all sequences of transformations.
Without being too formal, we state that this condition will hold
{\em generically} if all the transformations are chosen at random.
} 
Since a matrix with random entries will almost surely have
full rank, this leads to the following 
condition.
\begin{theorem}
Let a sequence of affine transformations $\m A_i: \R^{m_{i-1}} \rightarrow \R^{m_i}$,
for $i=1, \ldots, k-1$
be chosen at random.  Then $\m A_1 \ldots \m A_{k-1}$ almost surely has a right inverse if and
only if $m_i \ge m_0$ for $i=1, \ldots, k-1$.
\end{theorem}

\SKIP{
For practical purposes it is safe to proceed as if this result holds always (and not
just almost always), since the probability that $\m A_1\ldots \m A_{k-1}$ does not
have an inverse is vanishingly small (in fact zero).
}

Therefore, the strategy for selecting a set of parameters for a sequence of
affine layers may be described as follows.
For the sequence of layers to represent an
identity transform, we need that the input and output space have the same
dimension, namely $m_0 = m_k$.  Then
\begin{enumerate}
\vspace{-0.3\baselineskip}
\item Select intermediate dimensions $m_1, \ldots, m_{k-1}$ such that
$m_i \ge m_0$ for all $i=1, \ldots, k-1$.
\vspace{-0.3\baselineskip}
\item Define random affine transforms $\m A_i$ by selecting the entries of 
matrices $\mathbf A_i$ and vectors $\vv b_i$ randomly, using a suitable random number
generator, for instance by a zero-mean normal (Gaussian) distribution.
\vspace{-0.3\baselineskip}
\item Compute the composition $\m A_1 \m A_2 \ldots \m A_{k-1}$ by matrix multiplication,
and take its right-inverse.
\vspace{-0.3\baselineskip}
\item Set $\m A_k$ to equal $(\m A_1 \m A_2 \ldots \m A_{k-1})^R$.
\end{enumerate}
\vspace{-0.3\baselineskip}
The resulting composite mapping $\m A_1 \m A_2 \ldots \m A_k$ is the identity affine transformation.

\begin{figure}[t]
\centering
  \begin{subfigure}[b]{0.35\textwidth}
  \includegraphics[width=\textwidth]{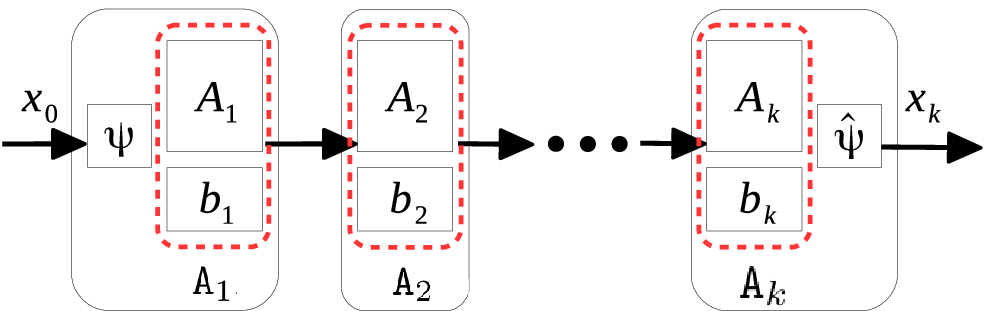}
  \vspace{-\baselineskip}
  \subcaption{without activation}
  \label{fig:T1_without}
  \vspace{-2mm}
  \end{subfigure}
  \begin{subfigure}[b]{0.48\textwidth}
  \includegraphics[width=\textwidth]{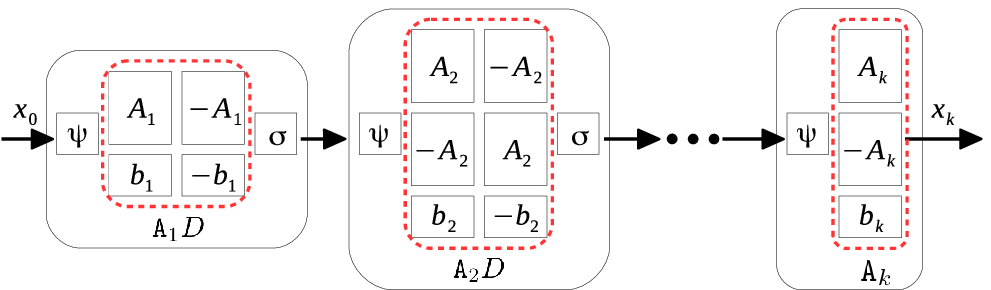}
  \vspace{-\baselineskip}
  \subcaption{with activation}
  \label{fig:T1_with}
  \end{subfigure}
\vspace{-\baselineskip}
\caption{\em\small Transparent initialization.
$\psi:\mathbf{x} \mapsto [\mathbf{x}, \mathbf{1}]$;
$\hat{\psi}:[\mathbf{x}, \mathbf{1}] \mapsto \mathbf{x}$;
$\sigma$: activation function.
The three modules, from left to right, in (b) are corresponding to Eqs.~\eqref{eq:first_init}-\eqref{eq:last_init}.}
\label{fig:TI}
\vspace{-3mm}
\end{figure}

\vspace{-1ex}
\paragraph{With Activation.}
Usually, each affine mapping $\m A$ will be followed by an
activation function $\sigma$.
For now, we assume that this is ReLU($\cdot$).
Our approach including activations is to apply the activation 
to both $\vv x$ and $-\vv x$ and then sum them.
Consider
\begin{equation}
\vv x \xmapsto{D}[\vv x, - \vv x] \xmapsto{\sigma} [\vv x^+, \vv x^-] \xmapsto{S} \vv x^+ - \vv x^- = \vv x ~,
\end{equation}
where mappings $D$ (duplicate) and $S$ (subtract) are the mappings
as shown above, and $\vv x^+$ and $\vv x^-$ are
the positive and negative components of $\vv x$.
This shows that $\vv x D\sigma S = \vv x$,
and so $D \sigma S$ is the identity mapping.
Note also that both $D$ and $S$ are affine (in fact linear) transformations.

Applying this to a sequence of affine transformations $\m A_i$
such that $\vv x \m A_1 \m A_2 \ldots \m A_k = \vv x$ 
gives
\begin{equation}
\vv x\, (\m A_1 D\sigma S)\, (\m A_2 D\sigma S)\, \ldots\, (\m A_{k-1} D\sigma S) \, \m A_k = \vv x ~.
\end{equation}
Bracketing this differently gives
\begin{equation}
\vv x (\m A_1 D \sigma)\, (S \m A_2 D \sigma)\, \ldots\,
(S \m A_{k-1} D \sigma)\, (S \m A_k) = \vv x ~,
\end{equation}
where now each bracket is an affine mapping followed by
the activation $\sigma$ (except the last).
This may then be  implemented as a sequence of affine layers
(convolution or fully-connected) with activations.
Thus, $\m A_1 D$ is the mapping
\begin{equation}
\label{eq:first_init}
\vv x \mapsto 
 \vv x  \mbegin{rr}
    \mathbf{A}_1 & -\mathbf{A}_1 
     \mend
+ [\mathbf b_1, -\mathbf b_1] ~.
\end{equation}
The mapping $S \m A_i D$ is the affine transformation
\begin{equation}
\label{eq:middle_init}
[\vv x^+, \vv x^-] \mapsto 
  [\vv x^+, \vv x^-]  \mbegin{rr}
    \mathbf{A}_i & -\mathbf{A}_i \\
     -\mathbf{A}_i & \mathbf{A}_i 
     \mend
+ [\mathbf b_i, -\mathbf b_i]
\end{equation}
and
$S \m A_k$ is the mapping
\begin{equation}
\label{eq:last_init}
[\vv x^+, \vv x^-] \mapsto 
  [\vv x^+, \vv x^-] \mbegin{r}
    \mathbf{A}_k\\
     -\mathbf{A}_k
     \mend
+ \mathbf b_k ~.
\end{equation}
\ifcomment
\zw{previously it was}
\begin{equation}
[\vv x^+, \vv x^-] \mapsto 
  [\vv x^+, \vv x^-] \mbegin{r}
    \mathbf{A}_k\\
     -\mathbf{A}_k
     \mend
+ [\mathbf b_k, -\mathbf b_k] ~. \nonumber
\end{equation}
\fi
The structure of the network may be represented as in Fig.~\ref{fig:TI}.
Observe that the outputs of intermediate layers have twice
the dimension of the output of the affine mappings $\m A_i$.

It is important to note that the structured form of the affine
mappings in Eqs.~\eqref{eq:first_init}-\eqref{eq:middle_init} are for initialization only.
There is
no sharing parameters (such as $\m A_i$ and $- \m A_i$)
and layers are free to implement any affine
transform during training.
In fact, $\mathbf x$ in any activation functions satisfying
\begin{equation}
\sigma(\mathbf x) - \sigma(-\mathbf x) = c\mathbf x,
\end{equation}
where $c$ is a non-zero constant,
can be recovered, such as SoftReLU($\cdot$) defined by
$\sigma(\mathbf x) = \log (1 + e^{\mathbf x})$ and LeakyReLU($\cdot$) \cite{leakyrelu}:
\begin{equation}
\label{eq:leaklyrelu}
\mathbf{x} = \frac{1}{1+\delta}\, \left[\sigma(\mathbf{x}),\, \sigma(-\mathbf{x}) \right]
    \begin{bmatrix}
    \mathbf{I} \\
    -\mathbf{I}
    \end{bmatrix},
\end{equation}
where $\delta>0$ is the slope of the negative part of LeakyReLU($\cdot$).
The substitution of Eq.~\eqref{eq:leaklyrelu} to
Eqs.~\eqref{eq:first_init}-\eqref{eq:middle_init} contributes to the
identity mapping in our transparent initialization.
One can derive it in a similar way to ReLU($\cdot$).
More details are given in the supplementary material.

\subsection{Logit Consistency with Sparse Encoder}

\begin{table}[t]
\centering
\resizebox{0.35\textwidth}{!}{
\begin{tabular}{cc|ccccccc}
\multicolumn{2}{c}{}&\multicolumn{7}{c}{pixel index} \\
&&0&1&2&3&4&$\cdots$&$N_p$-1 \\
\cline{2-9}
\multirow{5}{*}{\rotatebox[origin=c]{90}{superpixel index}}
&\rotatebox[origin=c]{90}{0}&0&1&1&0&0&$\cdots$&0 \\
&\rotatebox[origin=c]{90}{1}&0&0&0&1&1&$\cdots$&0 \\
&\rotatebox[origin=c]{90}{2}&1&0&0&0&0&$\cdots$&1 \\
&$\vdots$&$\vdots$&$\vdots$&$\vdots$&$\vdots$&$\vdots$&$\ddots$&$\vdots$ \\
&\rotatebox[origin=c]{90}{$N_s$-1}&0&0&0&0&0&$\cdots$&0 \\
\cline{2-9}
&sum&1&1&1&1&1&$\cdots$&1
\end{tabular}}
\caption{\em\small Example of sparse property for indexing $N_p$ pixels by
$N_s$ superpixels.
Each superpixel contains only a few pixels (``1" in each row)
for logit consistency.
Hence, an efficient encoding is achieved by a sparse 
$N_s \times N_p$ matrix with $N_p$ non-zero elements.}
\vspace{-5mm}
\label{tab:sparse_table}
\end{table}

In addition to the notations of pixel number $N_p$, superpixel
number $N_s$, and subsets of pixels
$\mathcal{P}=\{\mathcal{P}_0, ..., \mathcal{P}_{N_s-1}\}$ in
Sec.~\ref{sec:networks},
the label set is defined as $\mathcal{L}=\{0, ..., N_l-1\}$
given $N_l$ labels.
For logit $x^l_s$ of $\mathcal{P}_s$ for superpixel $s$ at label $l$,
the logit consistency follows
\begin{equation}
\label{eq:uniform}
x^{l}_{s}(p) \gets \frac{1}{|\mathcal{P}_s|}
\sum_{p \in \mathcal{P}_s} x^{l}_{s}(p),\ \forall p \in \mathcal{P}_s,
\end{equation}
which guarantees all pixels in $\mathcal{P}_s$ having
the same logit at each label so as to be assigned with the same label.

Nevertheless, considering the high complexity of indexing $x^l_s(p)$,
a dense matrix operation requires a large GPU memory,
that is $N_l N_s N_p$, especially for the back-propagation in CNN learning.
This makes it infeasible for training due to the limited GPU memory
in our experiments.

Hence, we adopted a sparse encoder, including sparse encoding and
decoding, with sparse matrix operations for the consistency.
Let us set matrix for indexing pixels by superpixels as $M(s,p)$,
logit matrix as $M(l,p)$, where $s \in \mathcal{S}$, $p \in \mathcal{P}$,
and $l \in \mathcal{L}$, sparse encoding and decoding are
\begin{subequations}
\label{eq:uniform_sparse}
\begin{align}
\text{Encoding:} \
M(s,l) &= \frac{\text{SMM}(M(s,p), M^T(l,p))}{\text{SADD}_{p \in
\mathcal{P}_s}(M(s,p))}, \\
\text{Decoding:} \
M(l,p) &\gets \text{SMM}(M^T(s,l), M(s,p)),
\end{align}
\end{subequations}
where $M(s,p) \in \mathcal{B}^{N_s N_p}$ is \{0, 1\} binary, shown in
Table~\ref{tab:sparse_table}, SMM($\cdot$) is sparse
matrix multiplication, and SADD($\cdot$) is sparse addition.
This converts Eq.~\eqref{eq:uniform} from dense operations to sparse
with reduced complexity from $N_l N_s N_p$ to $N_l N_p$.
Otherwise, it is infeasible to jointly train the networks due to the limited
GPU memory in our experiments.





\section{Experiments}
We first evaluated the properties of our transparent initialization
including its effectiveness of data recovery and numerical stability.
Then, we demonstrated its effect on jointly
learning pretrained networks of semantic segmentation and superpixels
together with a sparse encoder for logit consistency.
Our code in PyTorch will be released on GitHub upon publication.

\subsection{Properties of Transparent Initialization}

\paragraph{Effectiveness.}
Our transparent initialization aims at identically mapping
the output of linear layer(s) to the input at the early stage when
fine-tuning pretrained network(s).
It retains the effect of learned parameters of pretrained model(s).
Off-the-shelf parameter initialization methods, such
as random (uniform) and Xavier \cite{xavier} initialization, lead to
random values of the output at the early training stage,
which cannot generate effective features by the pretrained models.
Also, compared with the identical initialization with identity matrices
for deeper networks in Net2Net \cite{net2net}, our transparent
initialization has a high initialization rate with much more non-zero
parameters for dense gradients in the backpropagation.

We evaluated these methods on 3 fully-connected layers.
The (in\_channels, out\_channels) for each layer is
(42, 64)$\rightarrow$(64, 64)$\rightarrow$(64, 42).
Since Net2Net only supports square linear layer, \ie in\_channels
equals out\_channels, to increase network depth, all layers
have 42 in\_channels and out\_channels.
Input data is normally distributed with size (4, 42, 512, 512) as
(batch, in\_channels, height, width).

In Table~\ref{tb:ti_effect}, random and Xavier initialization have
$\sim$98\% initialization rate but cannot recover the output from its input,
leading to 0\% recovery rate.
Net2Net \cite{net2net} has only $\sim$2\% initialization rate and 50\%
recovery rate with
ReLU($\cdot$) for non-negative values only.
In contrast, our transparent initialization has a high initialization rate
and 100\% recovery rate by Eqs.~\eqref{eq:first_init}-\eqref{eq:last_init}
with ReLU($\cdot$).

\vspace{-1ex}
\paragraph{Numerical Stability.}
Since the effect of our transparent initialization is distributed
across layers and Eq.~\eqref{eq:right_inverse} is
achieved by a pseudo-inverse matrix for a rank-deficient matrix,
hidden layers have round-off errors, and thus, the matrix
multiplication of layer parameters is not strictly identical.
We therefore tested the numerical stability of our transparent
initialization on 4 numerical orders in Table~\ref{tb:ti_stable}.
Clearly, the max error between input and output is in proportion to
the magnitude of input values.
For a pretrained model, its output is usually stable in a numerical
range, such as probability in [0, 1].
One can easily enforce a numerical regularization if the model output is
out of range.

\begin{table}[t]
\centering
\setlength{\tabcolsep}{2pt}
\resizebox{0.48\textwidth}{!}{
\begin{tabular}{lrrrc}
\hline
\multicolumn{1}{c}{\multirow{2}{*}{Manner}}
&\multicolumn{1}{c}{\multirow{2}{*}{\makecell[c]{Init. \\Rate$\uparrow$}}}
&\multicolumn{2}{c}{Recovery Rate$\uparrow$}
&\multicolumn{1}{c}{\multirow{2}{*}{
\makecell[c]{Non-square\\Filter Supported}}}\\
&&w/o Activation& w Activation & \\
\hline
Random&98.2&0.0&0.0&\cmark \\
Xavier \cite{xavier}&98.2&0.0&0.0&\cmark \\
Net2Net \cite{net2net}&2.3&100.0&50.0&\xmark \\
Ours&\bf99.9&\bf100.0&\bf100.0&\cmark \\
\hline
\end{tabular}}
\vspace{-0.5\baselineskip}
\caption{\em\small \textit{Our transparent initialization has high initialization
and recovery rates on 3 fully-connected (FC) layers,
and supports non-square filters}.
``init. rate": percentage of non-zero (absolute value $>\epsilon$)
parameters;
``recovery rate": percentage of outputs with the same
(difference $<\epsilon$) values as inputs;
``activation": ReLU($\cdot$);
``non-square filter": a FC layer with 
different in\_channels and out\_channels.
Inputs are in [-10, 10] and
$\epsilon$=1e-4.
}
\label{tb:ti_effect}
\end{table}

\begin{table}[t]
\centering
\setlength{\tabcolsep}{5pt}
\resizebox{0.48\textwidth}{!}{
\begin{tabular}{rrrrr}
\hline
&\multicolumn{1}{c}{[-1, 1]}
&\multicolumn{1}{c}{[-10, 10]}
&\multicolumn{1}{c}{[-1e2, 1e2]}
&\multicolumn{1}{c}{[-1e3, 1e3]}\\
\hline
Max Error & $\sim$6.8e-6& $\sim$6.6e-5& $\sim$6.5e-4 & $\sim$6.6e-3 \\
\hline
\end{tabular}}
\vspace{-0.5\baselineskip}
\caption{\em\small Stability of transparent initialization
with numerical orders {1, 10, 1e2, 1e3}.
Layer parameters are consistent with Table~\ref{tb:ti_effect}.
}
\label{tb:ti_stable}
\vspace{-3mm}
\end{table}

\subsection{Implementation Setup}

\paragraph{Datasets.} \label{sec:dataset}
We evaluated our proposal on 3 popular semantic segmentation datasets,
ADE20K, PASCAL VOC 2012, and PASCAL Context.
\textit{ADE20K} \cite{ade20k_1,ade20k_2} has 150 semantic categories for
indoor and outdoor objects.
It contains 20,210 samples for training and 2,000 samples for validation.
\textit{PASCAL Context} \cite{pcontext} has additional annotations for
PASCAL VOC 2010 and provides annotations for the whole scene
with 400+ classes.
We selected the given 59 categories for semantic segmentation with 4,996
training samples and 5,104 validation samples.
\textit{PASCAL VOC 2012} \cite{voc2012} and Berkeley
benchmark \cite{berkeley} were used as a combined version for 21
classes segmentation.
This dataset has 1,449 images from PASCAL VOC 2012 val set for validation
and 10,582 images for training.
Meanwhile, \textit{MS-COCO} \cite{coco} was used to pretrain the semantic
segmentation network, \ie DeepLabV3+ in our case, for PASCAL VOC 2012.
It has 92,516 images for training and 3,899 images for validation, while 
20 object classes from the primary 80 classes were selected in accordance
with PASCAL VOC 2012.

For superpixel networks, we used the state-of-the-art
\textit{superpixel network with FCN} from
\cite{sp_fcn} that was pretrained on Berkeley Segmentation Data Set and
Benchmarks 500 (BSDS500) \cite{sbds500} containing 500 images with
handcrafted ground truth edges.

\vspace{-3ex}
\paragraph{Learning Details.}
For the ablation study on \textit{PASCAL VOC 2012}, we trained DeepLabV3+
from scratch on the combined dataset above with crop size $512^2$.
We set Learning Rate (LR) as 0.007 for ResNet and 0.07 for ASPP and
decoder.
To train on MS-COCO,
LR was set to 0.01 for ResNet and 0.1 for the others.
LR for superpixel network pretrained on BSDS500 was the same
as ResNet which was pretrained on ImageNet and is accessible
in PyTorch model zoo.
We used SGD \cite{sgd} with momentum 0.9, weight decay 5e-4, and
``poly" scheduler for 60 epochs.

For the joint training, we set LR as 1e-6 for pretrained DeepLabV3+
and superpixel with FCN.
LR for the transparent initialization module is 1e-7.
These LRs were fixed in the joint training for 20 epochs.

For \textit{ADE20K} and \textit{PASCAL Context}, we fine-tuned the state-of-the-art
pretrained ResNeSt101 and superpixel with FCN with crop size $480^2$.
LR for TI module is 1e-9, and 1e-6 for the others.
We fine-tuned for 100 epochs by SGD with momentum 0.9 and weight decay 5e-4.
Batch size was decreased from 16 to 8 due to the limited GPU memory.

\vspace{-3ex}
\paragraph{Metrics.} For semantic segmentation,
we used mean Intersection over Union (mIoU) and
pixel accuracy \cite{fcn}.
For segmentation edges, we
used performance ratio of true edges to false edges \cite{pr_metric} and F-measure, $2RP/(R+P)$ with edge recall rate $R$
and precision $P$ \cite{f_measure_1,f_measure_2}.

\subsection{Ablation Study}
The ablation study was on PASCAL VOC 2012.
We first reproduced DeepLabV3+ with ResNet101, resulting
in 78.85\% mIoU comparable to 78.43\%
\footnote{https://github.com/jfzhang95/pytorch-deeplab-xception.git}.
Applying superpixel network over it by logit consistency increased the
mIoU by 0.37\%.
We then adopted ResNet152 for a qualified baseline with 77.78\% on
the full size for evaluations.

Applying superpixel contours increased the mIoU from 77.78\% to 78.15\%,
based on which fine-tuning by our transparent initialization further
increased it by 0.79\% (1.16\% to DeepLabV3+).
Pretrained on MS-COCO for PASCAL VOC 2012, our proposal has 0.61\%
increase of mIoU.
Note that most edges in the ground truth
were neglected in the
evaluation, marked white in Fig.~\ref{fig:voc_viz}(d).

Again, our goal is to preserve sharp edges aligned with
object contours by superpixels.
Fig.~\ref{fig:voc_viz} vividly shows the enhanced object edges,
especially objects that are highly contrastive to the background,
such as birds and human heads.

\begin{table}[t]
\centering
\setlength{\tabcolsep}{2pt}
\resizebox{0.48\textwidth}{!}{
\begin{tabular}{lccccc}
  \hline
  \multicolumn{1}{c}{Manner} & Backbone & SP & TI & MS-COCO
  & mIoU ($512^2$/full) \\
  \hline
  DeepLabV3+ \cite{deeplabv3+}&ResNet101&-&-&-&78.85 / 76.47\\
  \multicolumn{1}{l}{\multirow{1}{*}{\rotatebox{0}{Ours}}}
  &ResNet101&\cmark&-&-&\bf79.22 / \bf76.98\\
  \hline
  DeepLabV3+ \cite{deeplabv3+}&ResNet152&-&-&-&79.32 / 77.78\\
  \multicolumn{1}{l}{\multirow{2}{*}{\rotatebox{0}{Ours}}}
  &ResNet152&\cmark&-&-&79.94 / 78.15\\
  &ResNet152&\cmark&\cmark&-&\bf80.46 / \bf78.94\\
  \hline
  DeepLabV3+ \cite{deeplabv3+}&ResNet152&-&-&\cmark&82.62 / 80.76\\
  \multicolumn{1}{l}{\multirow{1}{*}{\rotatebox{0}{Ours}}}
  &ResNet152&\cmark&\cmark&\cmark&\bf83.39 / \bf81.37\\
  \hline
\end{tabular}}
\vspace{-0.5\baselineskip}
\caption{\em\small Ablation study: \textbf{single-scale} evaluation on PASCAL VOC 2012.
Ours used DeepLabV3+ with ResNet101 and ResNet152
and superpixel net with a 3-layer TI module.
``SP": superpixel with logit consistency;
``TI": transparent initialization.
``mIoU" is on $512^2$ and full image size.
}
\label{tb:mIoU}
\vspace{-3mm}
\end{table}

\begin{table}[t]
\centering
\resizebox{0.32\textwidth}{!}{
\begin{tabular}{lcc}
  \hline
  \multicolumn{1}{c}{Manner}
  & \multicolumn{1}{c}{pixAcc.}
  & \multicolumn{1}{c}{mIoU} \\
  \hline
    PSPNet \cite{psp} & 81.39 & 43.29 \\
    EncNet \cite{encnet} & 81.69 & 44.65 \\
    ResNeSt50 \cite{resnest} & 81.17 & 45.12 \\
    ResNeSt101 \cite{resnest} & \underline{82.07} & \underline{46.91} \\
    Ours\footnotemark[3] & \bf{82.37} & \bf{47.42} \\
  \hline
\end{tabular}}
\vspace{-0.5\baselineskip}
\caption{\em\small \textbf{Multiscale} evaluation on ADE20K.
Ours used ResNeSt101 and superpixel
with a 2-layer TI module.}
\label{tb:mIoU_ade20k}
\vspace{-3mm}
\end{table}

\begin{table}[t]
\centering
\resizebox{0.32\textwidth}{!}{
\begin{tabular}{lcc}
  \hline
  \multicolumn{1}{c}{Manner}
  & \multicolumn{1}{c}{pixAcc.}
  & \multicolumn{1}{c}{mIoU} \\
  \hline
    FCN (ResNet50) \cite{fcn} & 73.40 & 41.00 \\
    EncNet \cite{encnet} & 80.70 & 54.10 \\
    ResNeSt50 \cite{resnest} & 80.41 & 53.19 \\
    ResNeSt101 \cite{resnest} & \underline{81.91} & \underline{56.49}  \\
    Ours\footnotemark[3] & \bf{82.43} & \bf{57.32} \\
  \hline
\end{tabular}}
\vspace{-0.5\baselineskip}
\caption{\em\small \textbf{Multiscale} evaluation on PASCAL Context.
Ours used ResNeSt101 and superpixel
with a 3-layer TI module.
}
\label{tb:mIoU_pcontext}
\vspace{-5mm}
\end{table}
\footnotetext[3]{Since \cite{resnest} has 64 GPUs for state-of-the-art ResNeSt training
while only 4 GPUs are accessible for ours,
we used ResNeSt101 as baseline.}

\begin{figure}[t]
\begin{center}
\includegraphics[width=0.48\textwidth]{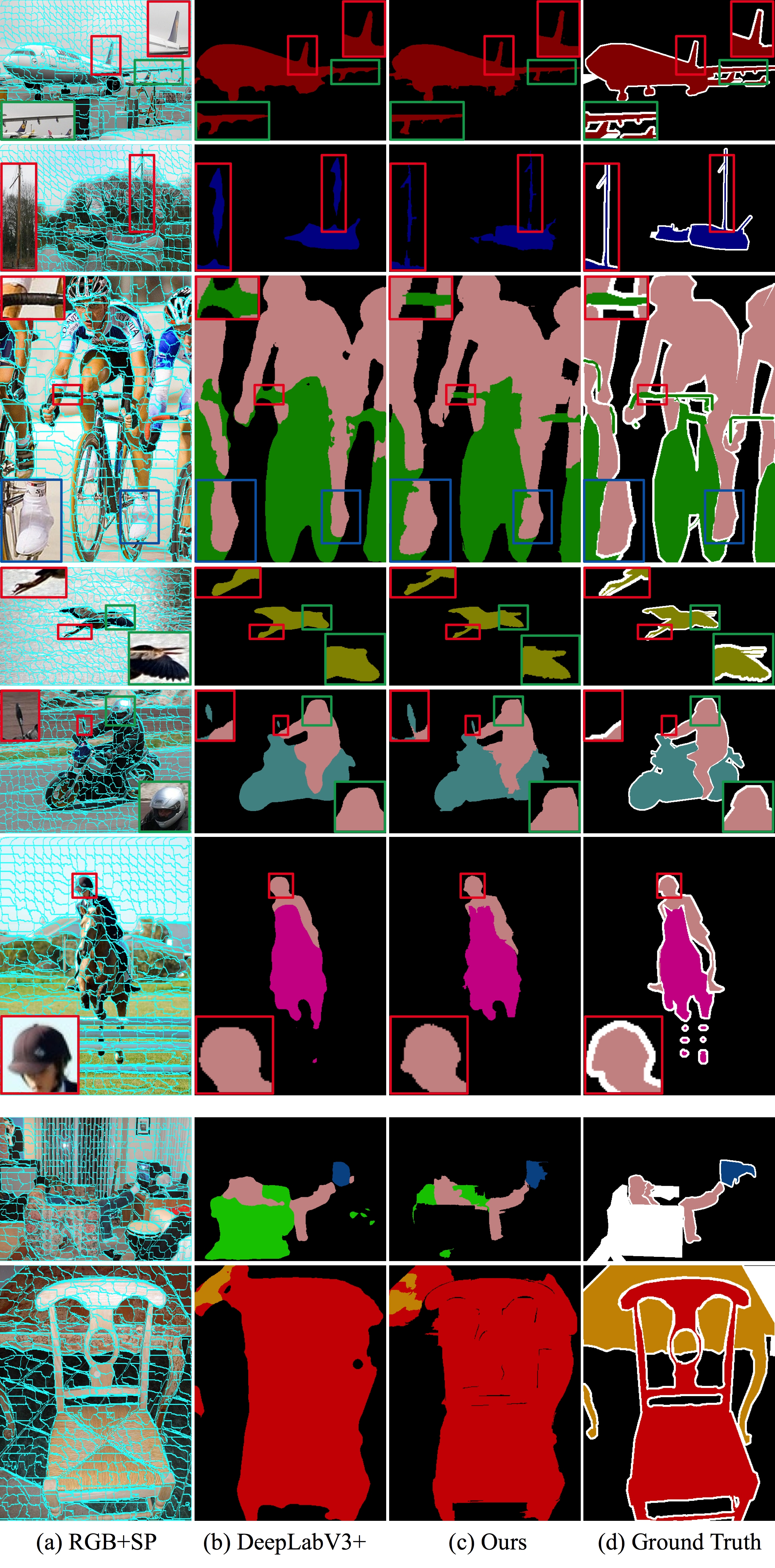}
\vspace{-1.8\baselineskip}
\caption{\em\small \textbf{Single-scale} evaluation on PASCAL VOC 2012.
First 6 rows are successful cases;
last 2 rows are failed cases.
SP maps are single-scale.
Best view by zoom-in.}
\label{fig:voc_viz}
\end{center}
\vspace{-10mm}
\end{figure}

\subsection{Evaluations}
\paragraph{Semantic Segmentation.}
For ADE20K and PASCAL Context, we used the most recent state-of-the-art 
semantic segmentation network ResNeSt \cite{resnest}
{\footnote{https://github.com/zhanghang1989/PyTorch-Encoding}}.
Its state-of-the-art performance on ADE20K using \textbf{ResNeSt200} is 82.45\% pixel accuracy and 48.36\% mIoU, and 83.06\% pixel accuracy and 58.92\% mIoU on PASCAL Context using \textbf{ResNeSt269}.
Since we have only 4 P100 (16 GB) GPUs for our experiments while \cite{resnest} has 64 V100 (16 GB) GPUs,
we chose ResNeSt101 instead of ResNeSt200 or ResNeSt269
as our baseline network.
Note that it is possible to enhance semantic
segmentation edges on those state-of-the-art networks given
sufficient GPU memory.

In Table~\ref{tb:mIoU_ade20k}, our method improves the pixel accuracy
by 0.39\% and mIoU by 0.61\% on ADE20K over the baseline.
In Table~\ref{tb:mIoU_pcontext}, the pixel accuracy is improved by
0.71\% and the mIoU by 0.83\% over the baseline.

More importantly, visualizations in Fig.~\ref{fig:ade20k_viz} for ADE20K and Fig.~\ref{fig:pcontext_viz} for PASCAL Context vividly show
the enhanced object edge details.
This is the core of our refinement on semantic segmentation with superpixel constraints.
For a fair comparison with \cite{resnest}, multiscale evaluations with multiscale superpixel maps were used
while the superpixel maps in
Figs.~\ref{fig:ade20k_viz}-\ref{fig:pcontext_viz} are single-scale
merely for demonstration.


\begin{figure}[t]
\begin{center}
\includegraphics[width=0.48\textwidth]{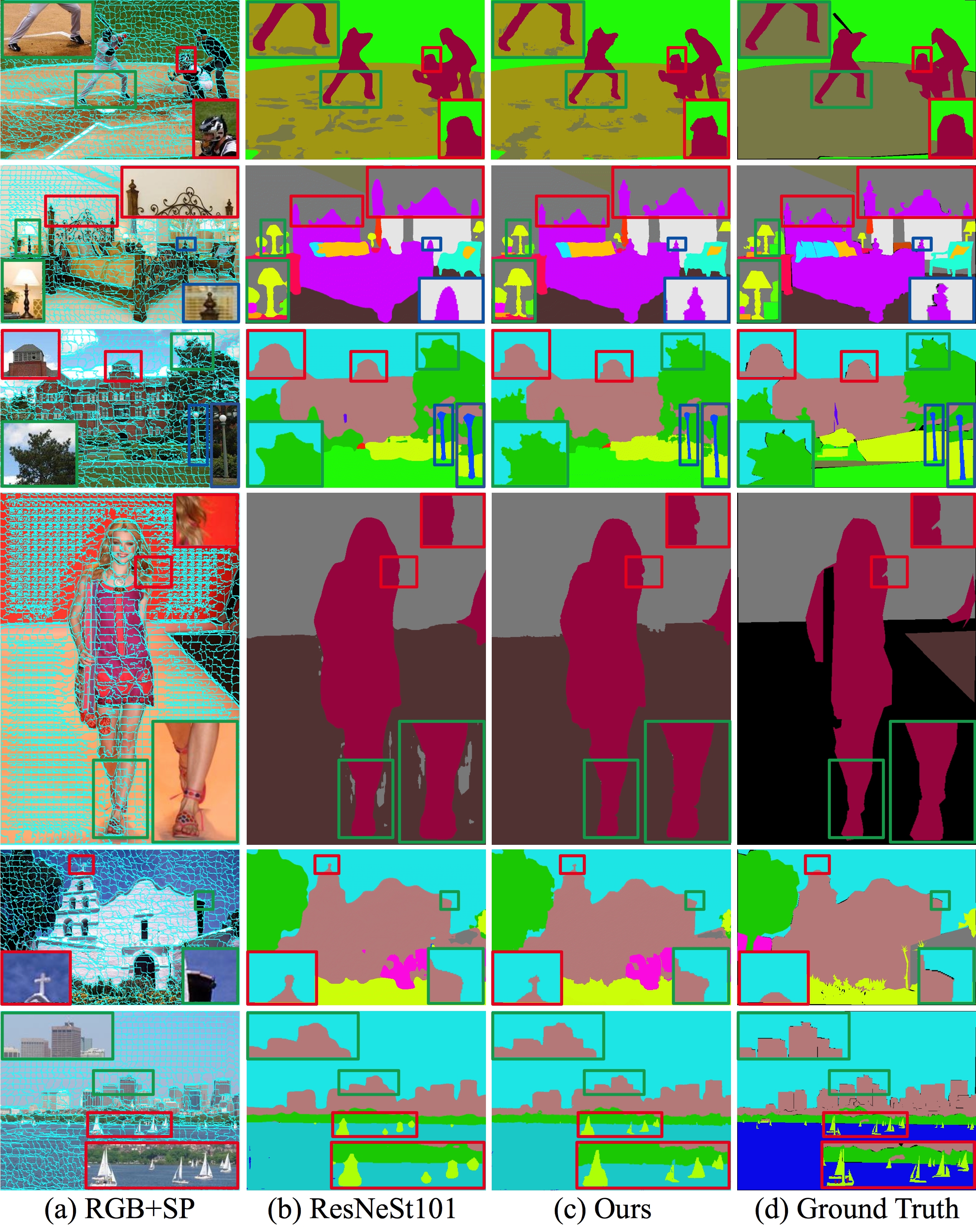}
\vspace{-1.5\baselineskip}
\caption{\em\small \textbf{Multiscale} evaluation on ADE20K.
SP maps are single-scale for demonstration.
Best view by zoom-in.}
\label{fig:ade20k_viz}
\end{center}
\vspace{-5mm}
\end{figure}

\begin{figure}[!h]
\begin{center}
\includegraphics[width=0.48\textwidth]{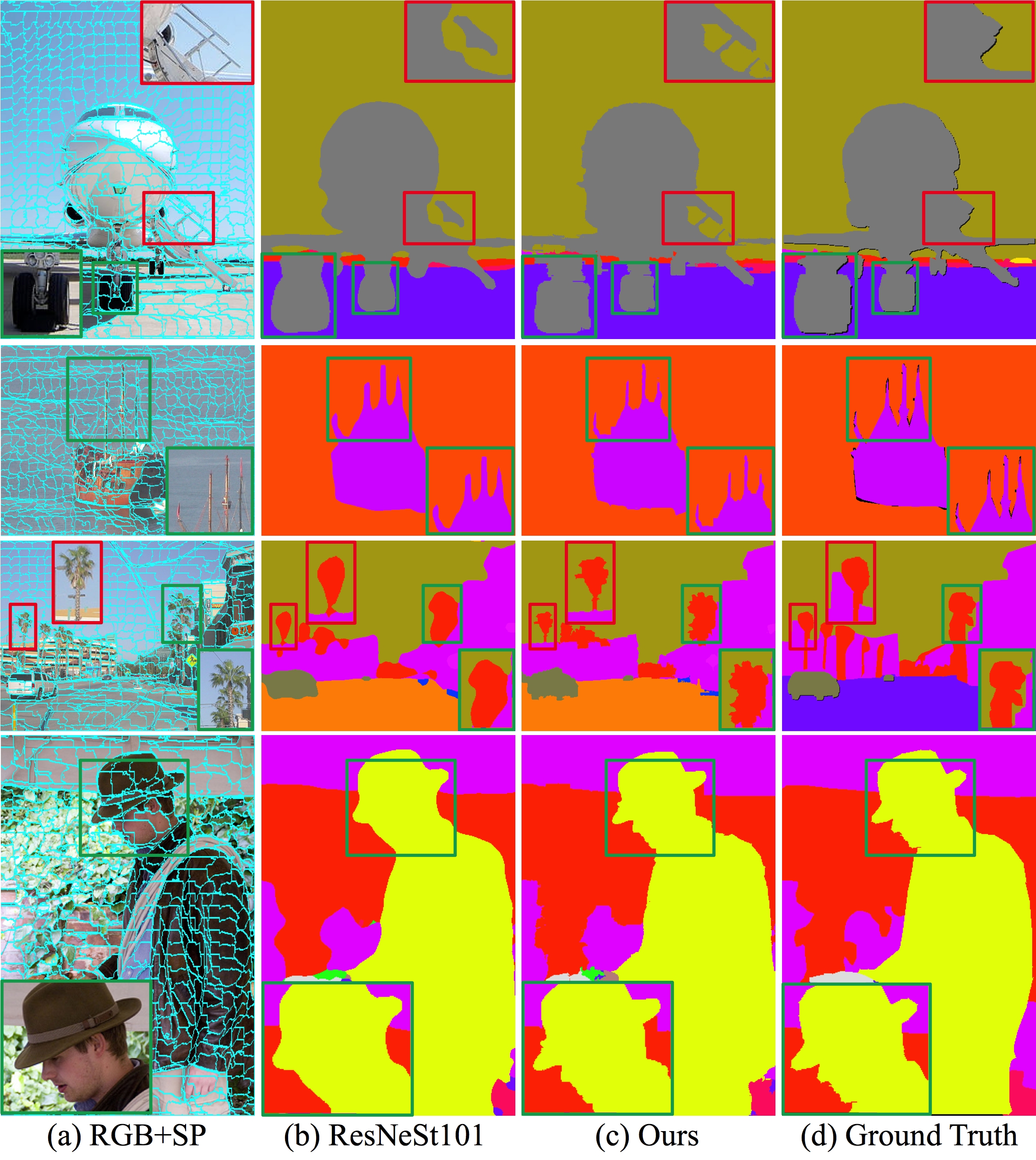}
\vspace{-1.5\baselineskip}
\caption{\em\small \textbf{Multiscale} evaluation on PASCAL Context.
SPs are single-scale for demonstration.
Best view by zoom-in.}
\label{fig:pcontext_viz}
\end{center}
\vspace{-7mm}
\end{figure}

\vspace{-3ex}
\paragraph{Semantic Segmentation Edges.}
Since our refinement of semantic segmentation mainly lies in object edge
areas, we used the popular Performance Ratio (PR) \cite{pr_metric}
and F-measure (FM) \cite{f_measure_1,f_measure_2} to evaluate the enhanced
segmentation edges.
In Fig.~\ref{fig:edge_curves}, ours outperform the baselines, \ie
ResNet101 on PASCAL VOC 2012 and ResNeSt101 on the others, as well as
the most recent state-of-the-art~\cite{resnest}, \ie
ResNeSt200 on ADE20K and ResNeSt269 on PASCAL Context, with higher
PR and FM.
Illustrations of the state-of-the-art edges compared with ours are shown in
Fig.~\ref{fig:demo}.

\begin{figure}[t]
\centering
\begin{subfigure}[b]{0.225\textwidth}
  \includegraphics[width=\textwidth]{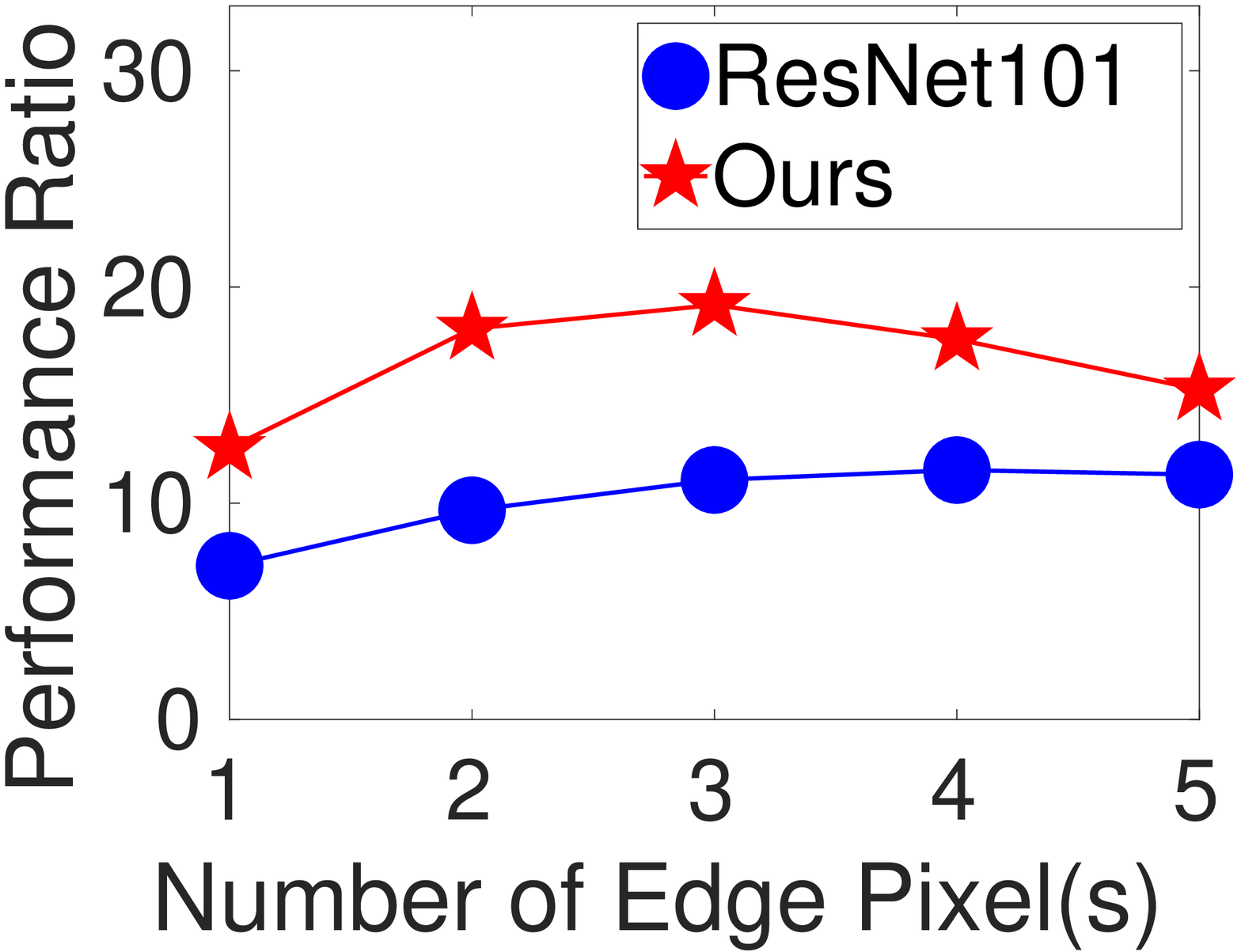}
  \subcaption{PR: PASCAL VOC 2012}
\end{subfigure}
\begin{subfigure}[b]{0.23\textwidth}
  \includegraphics[width=\textwidth]{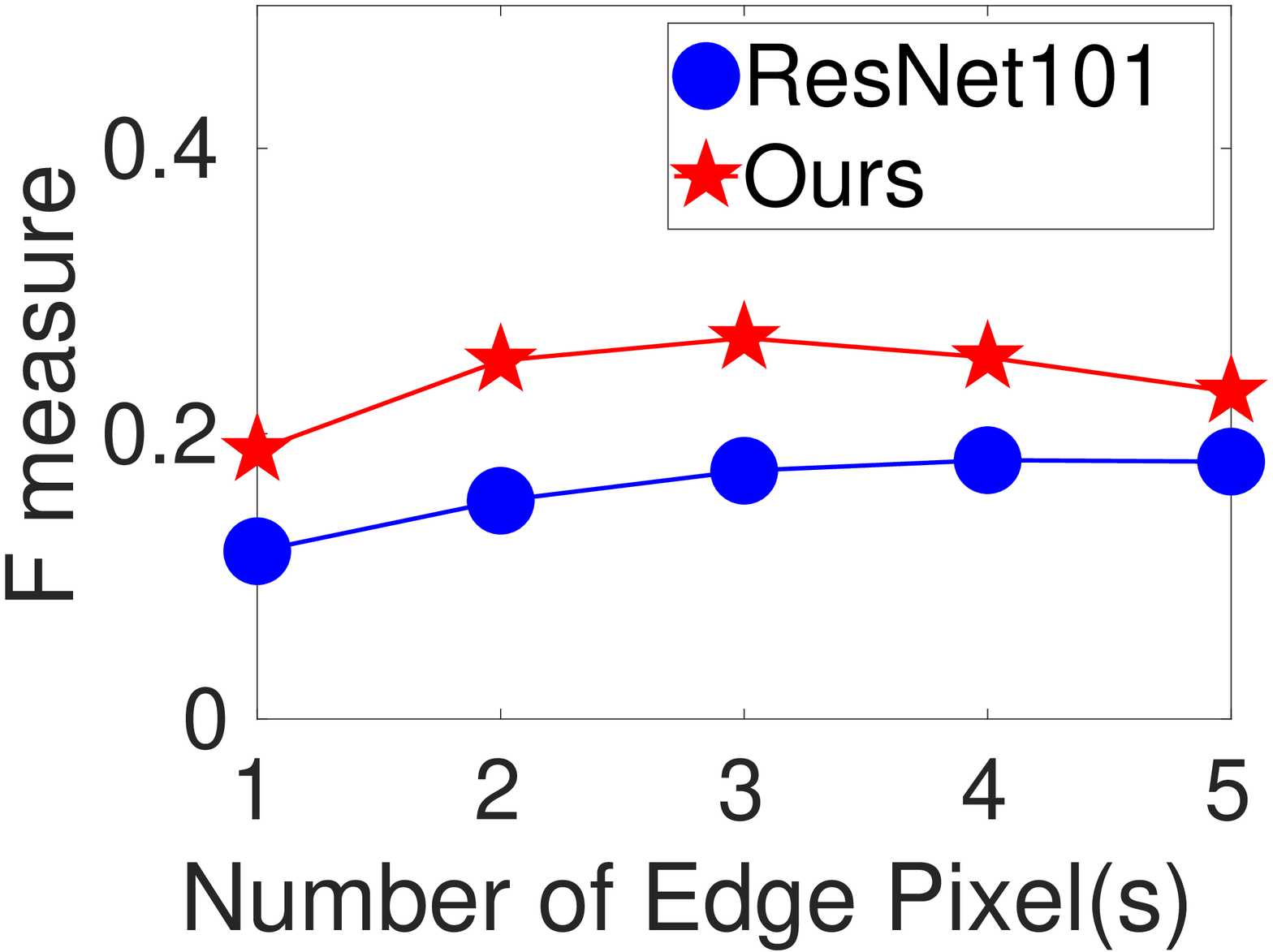}
  \subcaption{FM: PASCAL VOC 2012}
\end{subfigure}
\begin{subfigure}[b]{0.23\textwidth}
  \includegraphics[width=\textwidth]{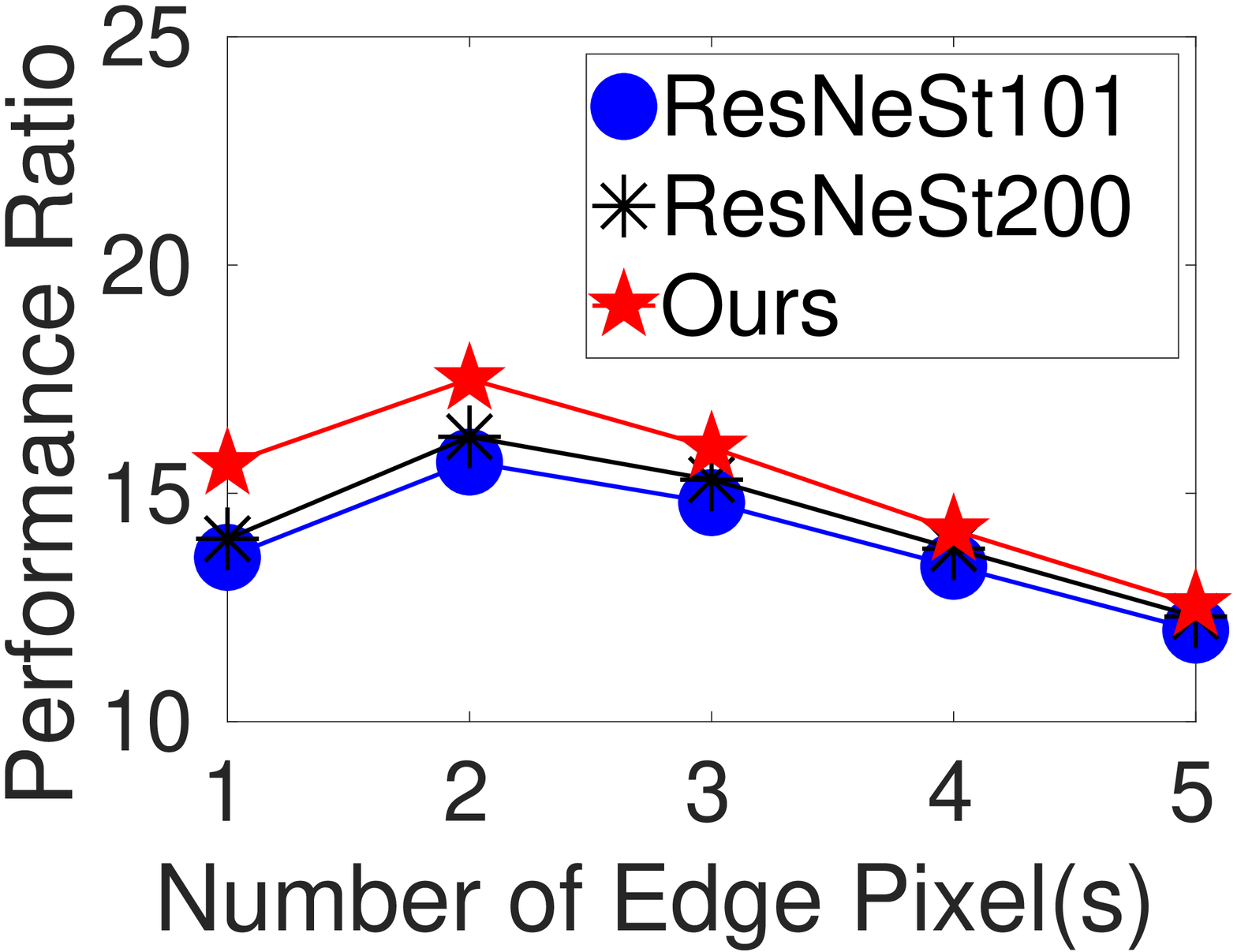}
  \subcaption{PR: ADE20K}
\end{subfigure}
\hspace{-3mm}
\begin{subfigure}[b]{0.238\textwidth}
  \includegraphics[width=\textwidth]{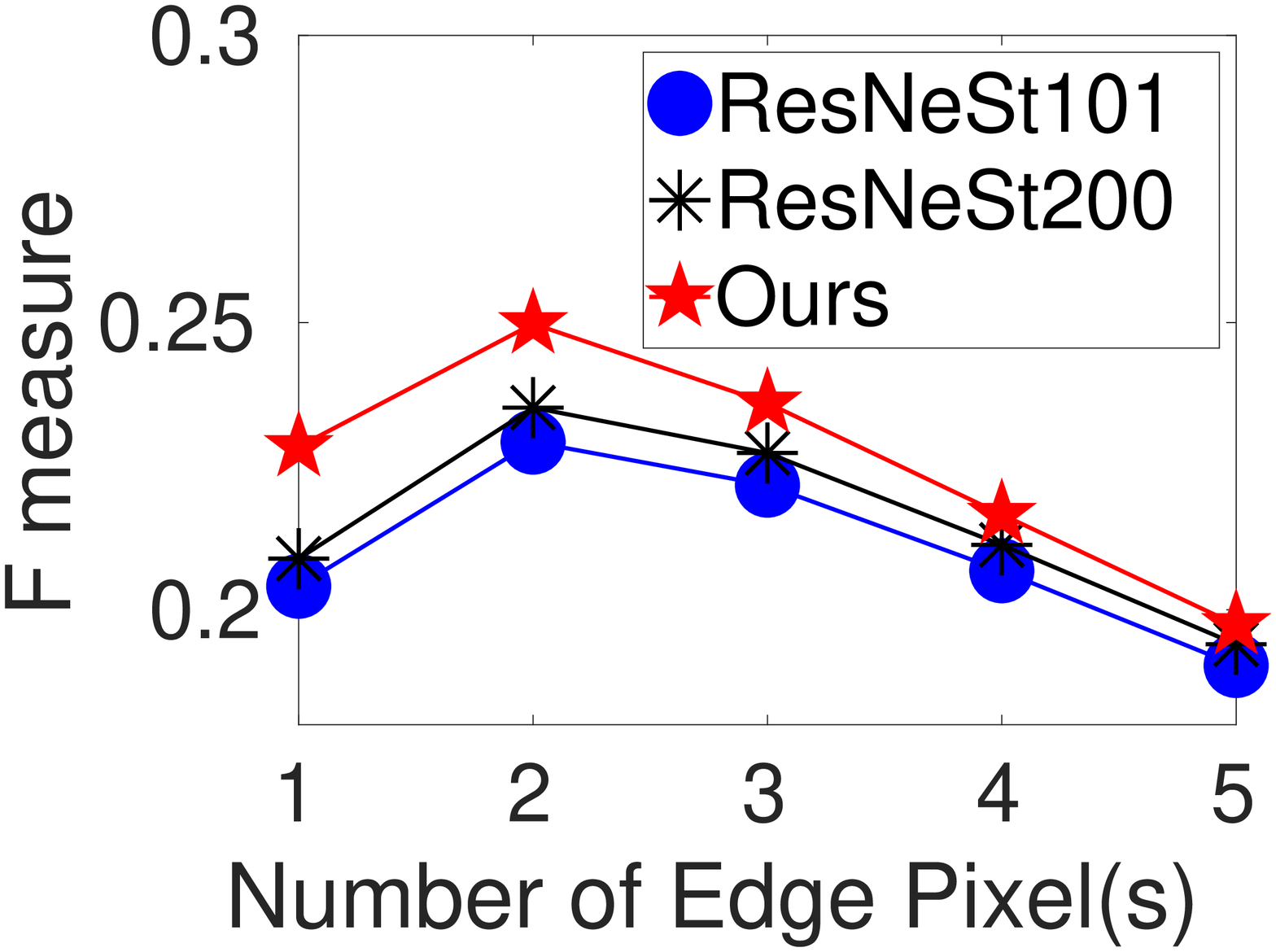}
  \subcaption{FM: ADE20K}
\end{subfigure}
\begin{subfigure}[b]{0.225\textwidth}
  \includegraphics[width=\textwidth]{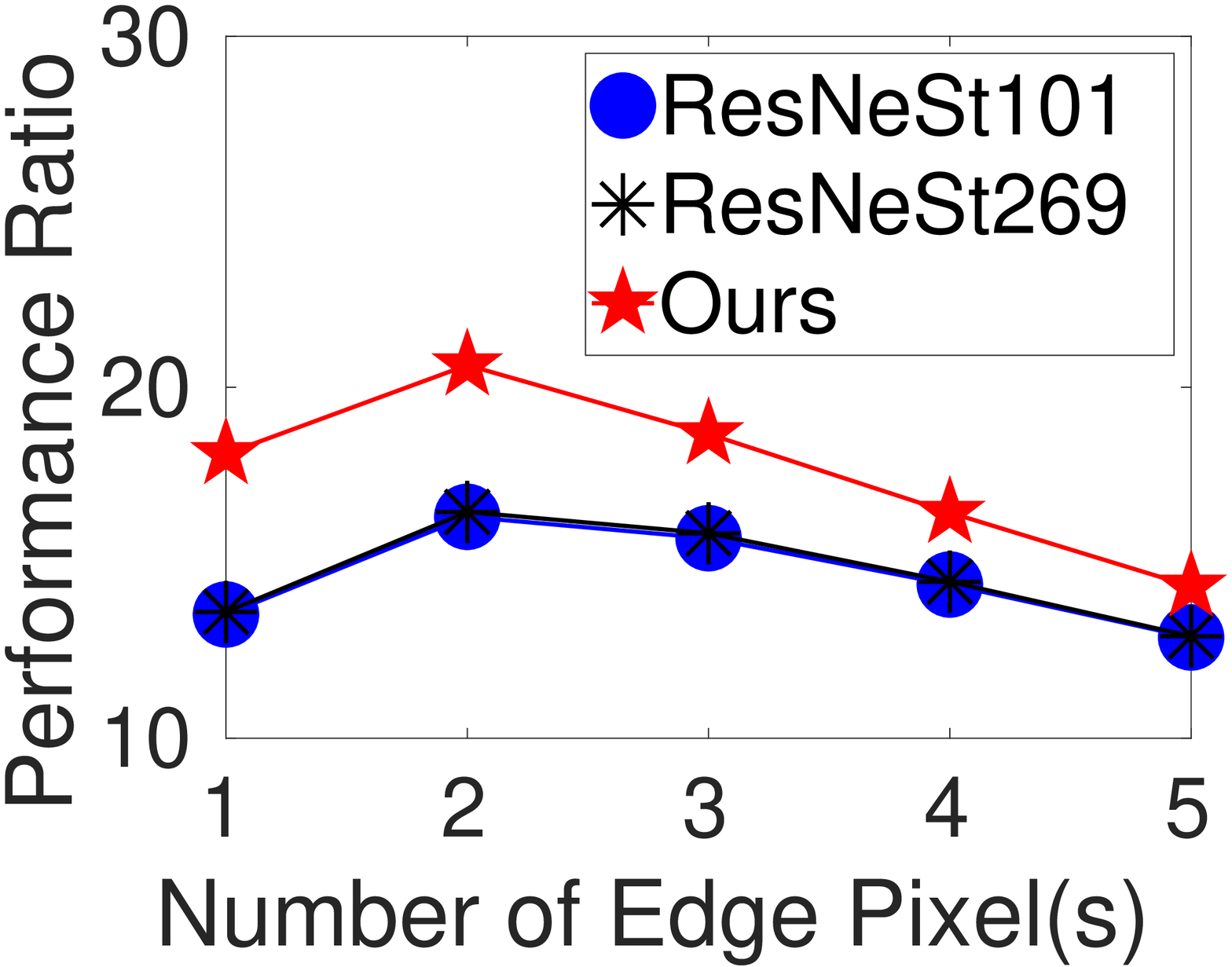}
  \subcaption{PR: PASCAL Context}
\end{subfigure}
\begin{subfigure}[b]{0.235\textwidth}
  \includegraphics[width=\textwidth]{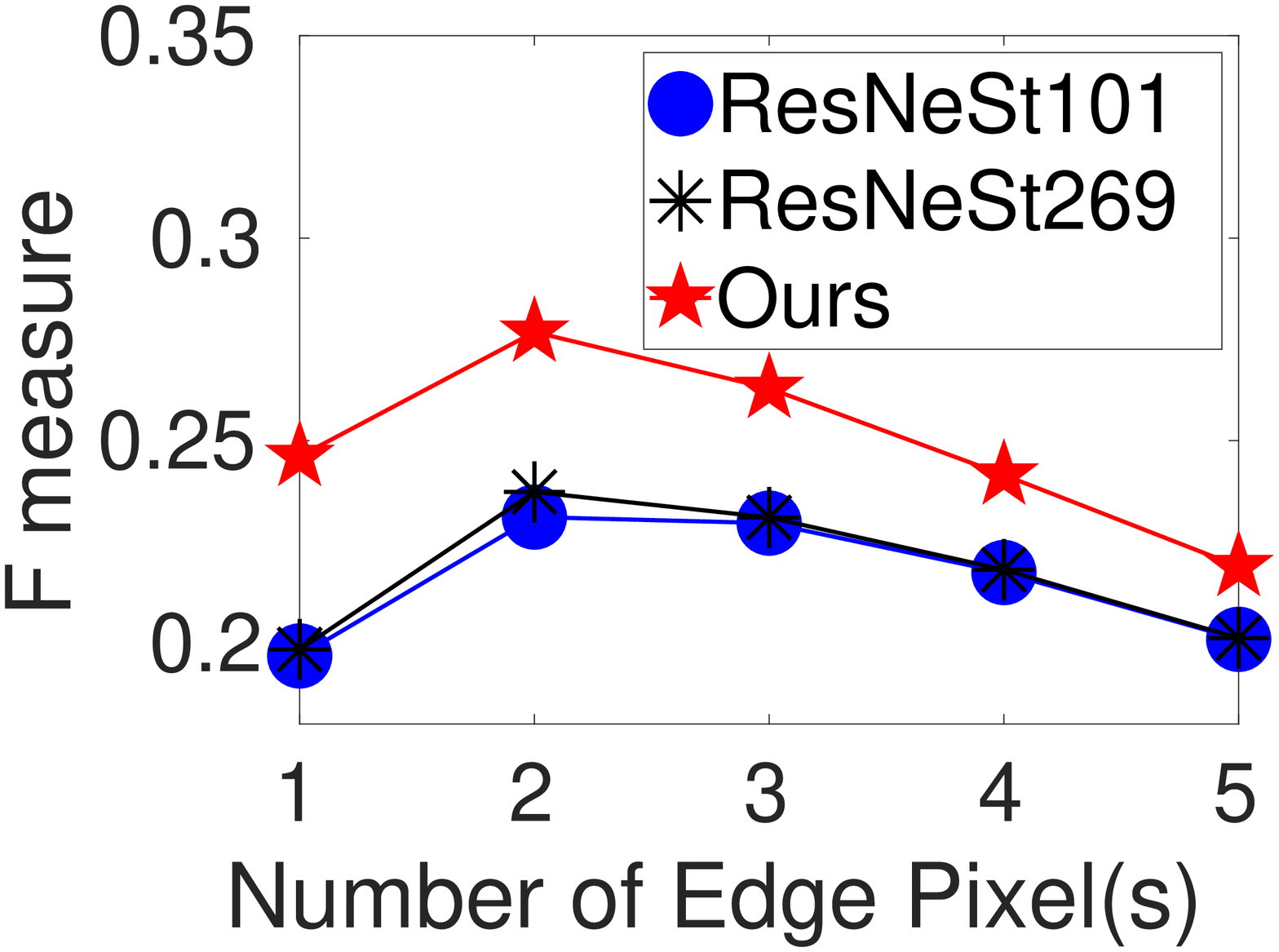}
  \subcaption{FM: PASCAL Context}
\end{subfigure}
\vspace{-0.5\baselineskip}
\caption{\em\small Evaluation on \textit{segmentation edges} using 
Performance Ratio (PR) and F-measure (FM).
Edges in ground truth are extended to \{1,2,3,4,5\} pixels.
Ours outperform the segmentation edges of the most recent state-of-the-art.
}
\label{fig:edge_curves}
\vspace{-7mm}
\end{figure}

\vspace{-1mm}
\section{Conclusion}
\vspace{-2mm}
With transparent initialization and sparse encoder introduced
in our paper, the joint learning of the state-of-the-art networks for
semantic segmentation and superpixels preserves object edges.
The proposed transparent initialization used to fine-tune pretrained
models retains the effect of learned parameters through identically
mapping the network output to its input at the early learning stage.
It is more robust and effective than other parameter initialization
methods, such as Xavier and Net2Net.
Moreover, the sparse encoder enables the feasibility of efficient
matrix multiplications with largely reduced computational complexity.
Evaluations on PASCAL VOC 2012, ADE20K, and PASCAL Context datasets
validate the effectiveness of
our proposal with enhanced object edges.
Meanwhile, the quality of our semantic segmentation edges, evaluated by
performance ratio and F-measure, is higher than other methods 
Additionally, transparent initialization is not limited to the joint
learning for semantic segmentation but can also be used to initialize
additional fully-connected layers for other tasks such as deep
knowledge transfer.
    
    \ifarxiv
      \appendix
      \subsection*{Acknowledgement}
We would like to thank the Australian National University, the Australian Centre for Robotic Vision (CE140100016), and Data61, CSIRO, Australia, for supporting this work.
      \subsection*{\Large{\textbf{Appendix}}}
      
    \else
      \clearpage
    \fi
\fi

{
\small
\bibliographystyle{ieee_fullname}
\bibliography{main}
}
\end{document}